%% file: main.tex
\documentclass[pdflatex,sn-nature]{sn-jnl}%

\usepackage{graphicx}%
\usepackage{multirow}%
\usepackage{amsmath,amssymb,amsfonts}%
\usepackage{amsthm}%
\usepackage{mathrsfs}%
\usepackage[title]{appendix}%
\usepackage{xcolor}%
\usepackage{textcomp}%
\usepackage{manyfoot}%
\usepackage{booktabs}%
\usepackage{algorithm}%
\usepackage{algorithmicx}%
\usepackage{algpseudocode}%
\usepackage{listings}%
\usepackage{stfloats}
\usepackage{CJKutf8}
\usepackage{tabularx}
\usepackage{makecell}
\usepackage{array}
\usepackage{booktabs}
\usepackage{caption}
\usepackage{longtable}
\usepackage{placeins}

\usepackage{./style/trackchanges}

\newcolumntype{L}{>{\raggedright\arraybackslash}X}

\theoremstyle{thmstyleone}%

\theoremstyle{thmstyletwo}%

\theoremstyle{thmstylethree}%

\begin{document}
\begin{CJK}{UTF8}{gbsn}

\input{sections/title}

\input{sections/abstract}
\input{sections/introduction} 
\input{sections/results}

\input{sections/discussion}

\input{sections/methods}

\backmatter

\bmhead{Author contributions}
Yibo Qiu led overall technical development, system integration and technical validation, coordinating implementation across modules. Zan Huang led robotic platform construction, control system development and debugging, with Haoliang Ye, Ao Tian, Shu'ang Sun and Handi Liu contributing to the robotics work. Yilin Qiao led the design, organization, execution and validation of biological experiments. Ziyang Ling developed MASBO. Yi Wang led optical system development, with Mohan Yang, Jun'ao Song and Haoran Zhang contributing to the optical work. Zhengyang Lu developed protocol generation and protocol-to-Action conversion. Hangke Peng contributed to drug-repurposing research, and Yuqing Long, Zhenzhen Zhou, Yu Xiong and Yaxuan Han contributed to biological experiments. Kan Ding and Di Wu performed image analysis. Ronald X Xu contributed to experimental planning and project supervision. Mingzhai Sun led overall research planning and project coordination, aligning research directions and guiding study implementation.

\bmhead{Competing Interests}
The authors declare no competing interests.

\bmhead{Data availability}
Source data are provided with this paper. Source Data 1 contains protocol-generation and action-mapping results; Source Data 2 contains culture-optimization measurements and condition summaries; Source Data 3 contains drug-candidate evidence and ATP measurements.

\bmhead{Code availability}
The BioMARS source code, configuration files and analysis scripts used in this study are provided in the supplementary ZIP file.

\begingroup
\setlength{\bibsep}{0.5em}

\input{references.bbl}
\endgroup

\input{appendices/appendices}

\end{CJK}
\end{document}

%% file: sections/title.tex
\title{BioMARS 2.0: a robotic biologist for formulating and adapting biological experiments}

\author[1,2]{\fnm{Yibo} \sur{Qiu}}\email{alexandreqiu@mail.ustc.edu.cn}
\equalcont{These authors contributed equally to this work.}

\author[1,2]{\fnm{Zan} \sur{Huang}}\email{huangzan@mail.ustc.edu.cn}
\equalcont{These authors contributed equally to this work.}

\author[1,2]{\fnm{Yilin} \sur{Qiao}}\email{qiaoyilin1997@mail.ustc.edu.cn}
\equalcont{These authors contributed equally to this work.}

\author[1,2]{\fnm{Ziyang} \sur{Ling}}\email{zaneling@mail.ustc.edu.cn}

\author[1,2]{\fnm{Yi} \sur{Wang}}\email{wangyi\_01@mail.ustc.edu.cn}

\author[1,2]{\fnm{Zhengyang} \sur{Lu}}\email{leolululu@mail.ustc.edu.cn}

\author[1,2]{\fnm{Hangke} \sur{Peng}}\email{penghangke@gmail.com}

\author[1,2]{\fnm{Haoliang} \sur{Ye}}\email{yehubert233@mail.ustc.edu.cn}

\author[1,2]{\fnm{Ao} \sur{Tian}}\email{ta99@mail.ustc.edu.cn}

\author[1,2]{\fnm{Shu'ang} \sur{Sun}}\email{sunshuang@mail.ustc.edu.cn}

\author[1,2]{\fnm{Handi} \sur{Liu}}\email{lhd24916055@mail.ustc.edu.cn}

\author[1,2]{\fnm{Yuqing} \sur{Long}}\email{islongyq@163.com}

\author[1,2]{\fnm{Mohan} \sur{Yang}}\email{mhyang379@mail.ustc.edu.cn}

\author[1,2]{\fnm{Kan} \sur{Ding}}\email{dingkan@mail.ustc.edu.cn}

\author[1,2]{\fnm{Zhenzhen} \sur{Zhou}}\email{zzz25@ustc.edu.cn}

\author[1,2]{\fnm{Di} \sur{Wu}}\email{wdwu@mail.ustc.edu.cn}

\author[1,2]{\fnm{Yu} \sur{Xiong}}\email{xiongyu1209@mail.ustc.edu.cn}

\author[1,2]{\fnm{Yaxuan} \sur{Han}}\email{hanyaxuan@mail.ustc.edu.cn}

\author[1,2]{\fnm{Jun'ao} \sur{Song}}\email{songjunao@mail.ustc.edu.cn}

\author[1,2]{\fnm{Haoran} \sur{Zhang}}\email{zhr8599@mail.ustc.edu.cn}

\author*[1,2]{\fnm{Ronald X} \sur{Xu}}\email{xux@ustc.edu.cn}
\author*[1,2]{\fnm{Mingzhai} \sur{Sun}}\email{mingzhai@ustc.edu.cn}

\affil*[1]{\orgdiv{Suzhou Institute for Advanced Research}, \orgname{University of Science and Technology of China}, \orgaddress{\city{Suzhou}, \state{Jiangsu}, \country{China}}}
\affil*[2]{\orgdiv{School of Biomedical Engineering, Division of Life Sciences and Medicine}, \orgname{University of Science and Technology of China}, \orgaddress{\city{Hefei}, \state{Anhui}, \country{China}}}

%% file: sections/abstract.tex
\abstract{
Biological experimentation is not reducible to a single optimization problem, but requires decisions at multiple levels of abstraction: translating scientific knowledge into experimental procedures, adapting culture conditions to measured phenotypes and selecting interventions that test biological hypotheses. Automating experimental execution does not, by itself, determine which experiments should be performed next or how they should change in response to biological feedback. Here we introduce BioMARS, a human-in-the-loop multi-agent robotic system that connects knowledge-guided experimental decision-making with robotic execution and biological feedback. At the procedural level, BioMARS generated literature-grounded protocols and traceable action plans, improving expert-reviewed protocol quality scores across four model backends. At the parametric level, BioMARS used image-derived phenotypes from robotic retinal organoid culture to select subsequent experimental conditions, with higher reported scores in later rounds of the three-round experiment. At the intervention level, BioMARS nominated drug-repurposing candidates from biomedical evidence and advanced six candidates to wet-lab testing by researchers. Together, these studies demonstrate feedback-driven robotic biological experimentation within a framework that supports decisions about experimental procedures, culture conditions and biological interventions.
}

\maketitle

%% file: sections/introduction.tex
\section{Introduction}\label{sec:introduction}

Laboratory automation has made experimental execution increasingly reproducible and scalable \cite{dettinger2022open}. Extending the use of robotic systems in biological research \cite{kanda2022robotic,tristan2021robotic,lan2025autobio,hamm2024modular,koniger2024rebia} requires connecting this execution capacity with decisions about experimental procedures, culture conditions and biological interventions. Such decisions draw on scientific knowledge to formulate experiments and on measured biological responses to guide subsequent investigations.

Robot scientists and self-driving laboratories have shown that experimental measurements can be returned to algorithms to guide condition selection \cite{king2004robotscientist,burger2020mobile,macleod2020selfdriving,szymanski2023alab,kanda2022robotic}. In parallel, language-model-based scientific agents can retrieve literature and database knowledge, generate hypotheses, design experiments and interpret data \cite{odonoghue2023bioplanner,qu2025crisprgpt}; some have been coupled to laboratory automation for experimental hypothesis testing \cite{boiko2023autonomous,swanson2025virtuallab,gottweis2026coscientist,ghareeb2026robin,huang2026biomni,qiu2025biomars}. Building on these advances, we ask how knowledge-guided experimental decisions can be connected to robotic culture and phenotype feedback, so that measured biological responses inform subsequent experiments.

Retinal organoid research provides a demanding setting in which to examine this connection. Retinal organoids reproduce key features of retinal development and provide experimentally accessible models for developmental and disease studies, but their morphology and developmental state vary substantially between cultures \cite{eiraku2011opticcup,nakano2012retina,zhong2014retinal,cowan2020retina,capowski2018staging,brooks2019organoid,harkin2024organoid}. This variability motivates the use of phenotypic feedback to guide culture optimization. Their relevance to retinal development and disease also provides a setting for testing interventions nominated from biomedical evidence.

Here we introduce BioMARS (Biological Multi-Agent Robotic System), a human-in-the-loop multi-agent robotic system that connects knowledge-guided experimental decision-making with robotic execution and biological feedback. At the procedural level, BioMARS generates literature-grounded protocols and traceable action plans. At the parametric level, the system uses measured phenotypes from robotic culture to select subsequent conditions across three rounds of retinal organoid experiments. At the intervention level, the system nominates drug-repurposing candidates from biomedical evidence for experimental testing by researchers.

%% file: sections/results.tex
\section{Results}
\label{sec:results}

\subsection{Experimental decision-making across three levels}
\label{sec:results_system}

\begin{figure*}[!tp]
\centering
\makebox[\textwidth][c]{%
\begin{minipage}{158mm}
\includegraphics[width=\linewidth]{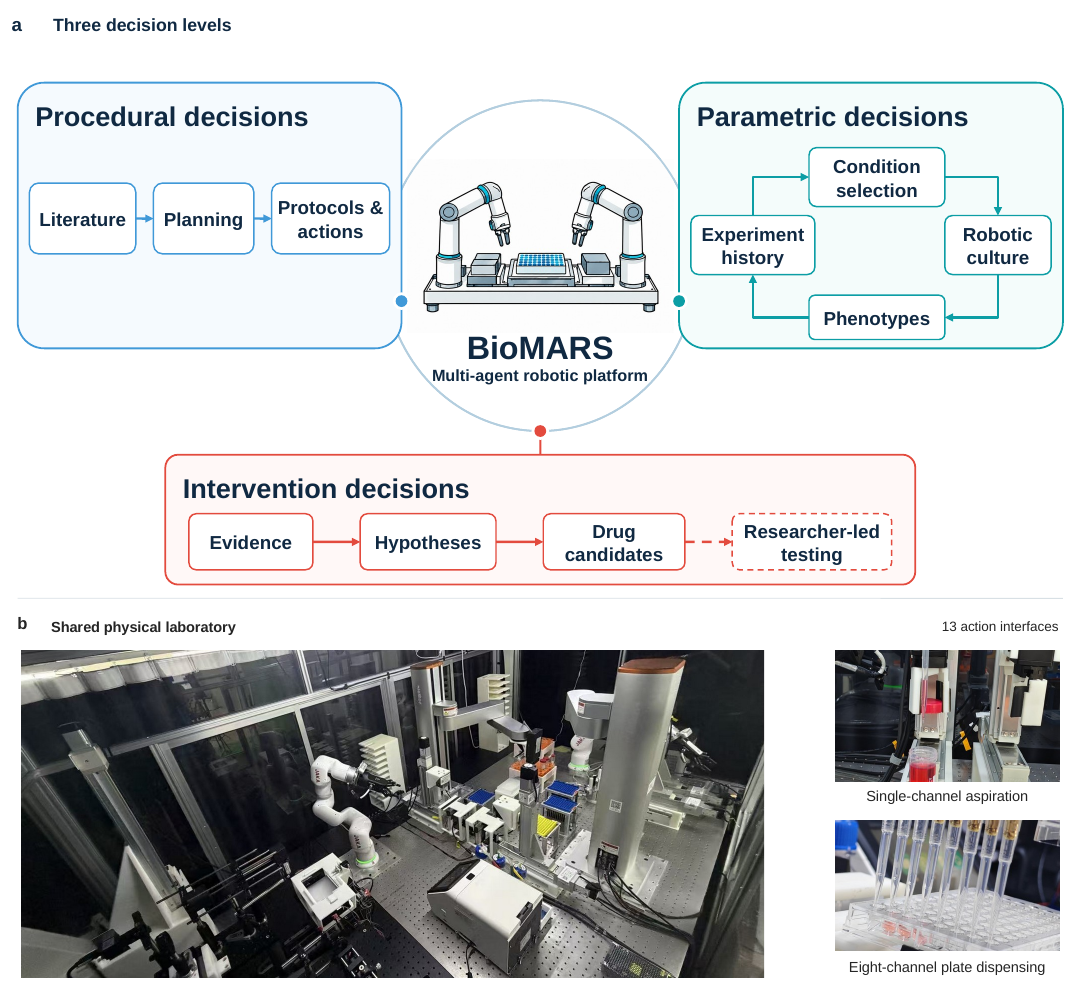}
\captionsetup{font=footnotesize}
\caption{\textbf{BioMARS across three levels of experimental decision-making.} \textbf{a}, Three tasks supported by a shared platform: protocol formulation, phenotype-guided culture optimization and evidence-guided intervention selection. Drug candidates undergo researcher-led testing. \textbf{b}, Automated cell-culture laboratory supporting 13 action interfaces, with single-channel aspiration and eight-channel plate dispensing shown at right.}
\label{fig:system}
\end{minipage}}
\end{figure*}

BioMARS supported three levels of experimental decision-making within a shared architecture: protocol formulation, culture-condition adaptation and intervention selection (Fig.~\ref{fig:system}). We evaluated these capabilities in three distinct tasks. The culture task linked condition selection, robotic execution and phenotypic feedback in a closed loop, with measured outcomes guiding subsequent experiments. The automated laboratory platform provided 13 action interfaces for material handling, liquid operations, culture and measurement (Table~\ref{tab:action_interfaces}). Generated protocols required human approval before laboratory use, and researchers performed drug testing.

\subsection{Procedural formulation from literature to action plans}
\label{sec:results_protocol_generation}

At the procedural level, BioMARS integrated operations, recipes and timing requirements from the literature into source-linked experimental plans before generating protocol text \cite{lewis2020rag,odonoghue2023bioplanner}. These plans specified experimental stages, step order and branch conditions, organizing dispersed source information into coherent workflows (Fig.~\ref{fig:protocol}a). In the mouse retinal differentiation task, BioMARS specified 5\% O$_2$ for days 0--10 and 20\% O$_2$ thereafter, together with complete medium replacement on day 26 followed by half-medium changes every two days. The corresponding FlashRAG output left these conditions and operations unspecified (Fig.~\ref{fig:protocol}b).

We compared BioMARS with FlashRAG \cite{jin2025flashrag} across 47 tasks and four model backends, evaluating 376 protocols. Under a five-dimensional rubric with expert-reviewed scores, BioMARS increased mean scores by 17.5\%, 20.9\%, 14.5\% and 17.8\% for GPT-4o, Qwen 3.6 Flash, DeepSeek V4 Flash and Gemini 3 Flash Preview, respectively (Fig.~\ref{fig:protocol}c), with higher scores on 40--44 tasks per backend (Table~\ref{tab:protocol_gains}). The largest differences were in content accuracy and procedural detail, which increased by 6.74--11.32 points out of 30 and 2.87--4.21 points out of 10, respectively (Fig.~\ref{fig:protocol_dimension_changes}). These gains reflected more complete retention of experimental content and operational parameters.

BioMARS connected protocol instructions to laboratory functions through explicit descriptions of the required operations \cite{jiang2024protocode}. Using step context, the system expanded experimental operations and associated handling, then assigned samples, containers and parameters to the corresponding calls (Fig.~\ref{fig:protocol}d,e). In the medium-change example, the brief instruction ``Medium is changed every other day'' became a sequence of calls to remove spent medium, deliver LIF + 2i medium from the designated reagent bottle and return the same dish to incubation. Container references maintained object continuity across these calls, turning the brief instruction into an action plan for a specific sample.

We evaluated how faithfully generated actions retained source requirements across 119 selected steps from 31 Blocks. Mean step-level matching scores were 85.56/100 with Qwen 3.6 Flash and 83.68/100 with DeepSeek V4 Flash (Fig.~\ref{fig:protocol}f).

\subsection{Parametric adaptation from measured phenotypes}
\label{sec:results_optimization}

\begin{figure*}[!tp]
\centering
\makebox[\textwidth][c]{%
\begin{minipage}{158mm}
\includegraphics[width=\linewidth]{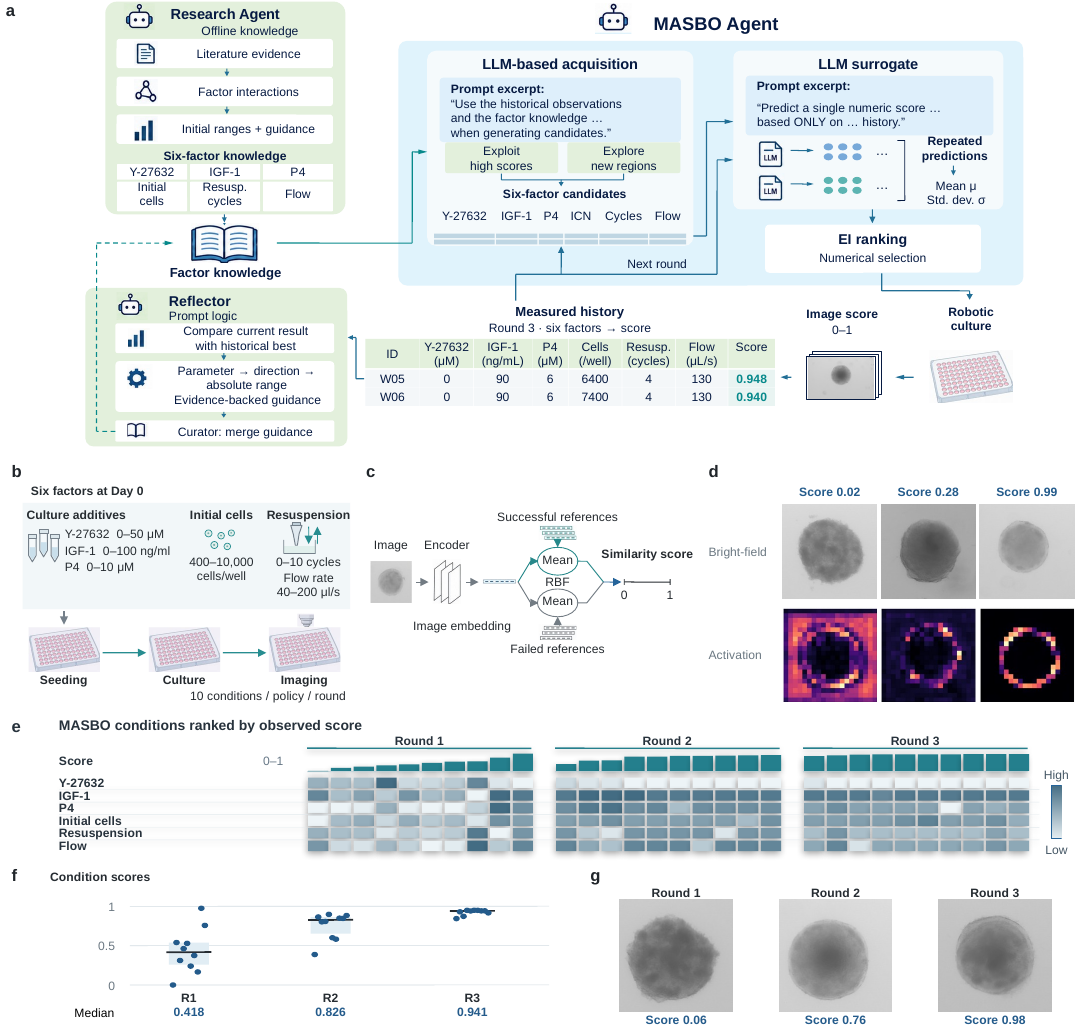}
\captionsetup{font=footnotesize}
\caption{\textbf{Image-derived feedback in robotic retinal organoid culture.} \textbf{a}, Knowledge- and phenotype-guided condition selection and robotic culture. Dashed links denote the implemented knowledge-update pathway. MASBO, multi-agent system--based Bayesian optimization. ICN, initial cell number; EI, expected improvement. \textbf{b}, Six-factor search space and culture workflow; ten conditions per method and round were tested on three plates. \textbf{c}, Image-based phenotype scoring from 0 to 1 using a radial basis function (RBF); higher scores indicate greater similarity to successful references. \textbf{d}, Bright-field images and layer-9 CNN responses with well-level scores. \textbf{e}, Factor combinations ranked by condition score, with colours scaled to the ranges in \textbf{b}. \textbf{f}, One three-round MASBO run ($n=10$ conditions per round). Dots denote conditions, boxes interquartile ranges, and horizontal lines and labels medians. Scores equally average condition means from the most highly correlated plate pair; recorded failures receive zero. \textbf{g}, Selected images from different wells across rounds, shown with equal-pixel crops and well-level scores.}
\label{fig:feedback}
\end{minipage}}
\end{figure*}

At the parametric level, we examined whether literature knowledge and feedback from completed cultures could guide condition selection for the next round of robotic experiments. Our multi-agent system--based Bayesian optimization (MASBO) approach varied six factors---Y-27632, IGF-1, progesterone (P4), initial cell number, resuspension cycles and flow rate---testing ten conditions in each of three rounds, with each condition measured on three plates (Fig.~\ref{fig:feedback}a,b). Offline literature retrieval provided factor-wise guidance for initial proposals. Once culture results were available, the system combined this guidance with measured history to generate candidates and ranked them by expected improvement calculated from repeated predictions by an LLM-based surrogate \cite{liu2024llambo}. Results from the first two rounds were returned before the next set of conditions was selected, allowing experimental feedback to inform subsequent choices.

To provide a common basis for comparing culture responses, we mapped bright-field images to a continuous phenotype score between 0 and 1, with higher scores indicating greater embedding similarity to human-labelled successful reference images (Fig.~\ref{fig:feedback}c). Plate scores were combined using the defined plate-concordance rule and served as the objective for subsequent condition selection.

Reported mean MASBO condition scores increased from 0.435 in round 1 to 0.752 in round 2 and 0.923 in round 3 (Fig.~\ref{fig:feedback}e,f). Seven of ten conditions scored at least 0.8 in round 2, and eight scored at least 0.9 in round 3.

Conventional Bayesian optimization (BO) also tested ten conditions per round over three rounds \cite{jones1998ego,shahriari2016bo,frazier2018bo}. In this three-round run, MASBO achieved higher mean condition scores in rounds 2 and 3. Complete comparisons, aggregation sensitivity analyses and retrospective Iris and iPSC benchmarks are provided in Appendix~\ref{app:culture_optimization}.

To distinguish responses among condition combinations within the high-scoring region, we compared individual conditions tested in round 3. An early high-scoring formulation scored 0.873 when retested, whereas a combination with the same initial cell number, Y-27632 and IGF-1 settings but altered P4, resuspension cycles and flow rate scored 0.948, with higher scores on all three plates. Another pair of conditions in the same round differed only in Y-27632 concentration: 0 and 5~$\mu$M yielded scores of 0.948 and 0.843, respectively. These local comparisons resolved differences in response within the high-scoring region, providing experimental evidence to guide joint adjustment of culture formulations and robotic operating parameters (Fig.~\ref{fig:feedback}e; Source Data 2).

\clearpage
\newgeometry{top=19mm,bottom=23mm,left=19mm,right=19mm,bindingoffset=0mm}
\begin{figure*}[p]
\centering
\includegraphics[width=172mm]{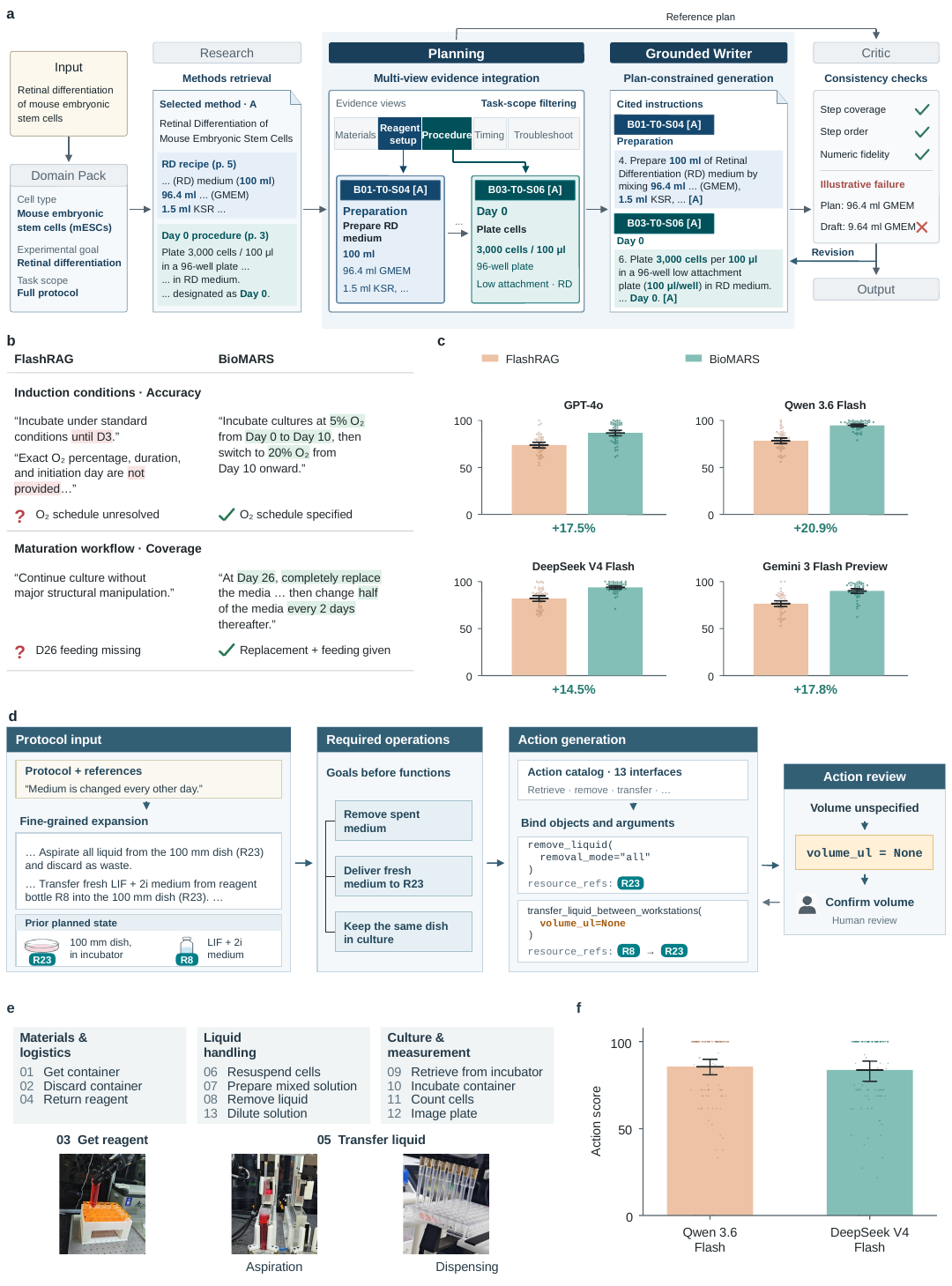}
\captionsetup{font=footnotesize,skip=3pt}
\caption{\textbf{Procedural-level formulation and action mapping.} Full legend follows.}
\label{fig:protocol}
\end{figure*}
\clearpage
\restoregeometry
\begin{figure*}[!t]
\ContinuedFloat
\centering
\makebox[\textwidth][c]{%
\begin{minipage}{158mm}
\captionsetup{font=footnotesize,skip=3pt}

\caption{\textbf{Procedural-level formulation and action mapping.} \textbf{a}, Source-linked protocol generation illustrated by retinal differentiation medium preparation and day-0 cell plating. Selected steps are shown. Source A, La Torre, \textit{Bio-protocol} \textbf{6}, e1851 (2016) \cite{latorre2016retinal}. \textbf{b}, Qwen 3.6 Flash output excerpts comparing oxygen conditions and medium-replacement schedules under FlashRAG and BioMARS. Question marks indicate unresolved information; checks indicate information provided; ellipses mark omissions. \textbf{c}, Expert-reviewed protocol scores across four models (47 paired tasks per model; 376 protocols). Percentages indicate relative mean gains over FlashRAG. \textbf{d}, Mapping expanded medium-change operations to functions and parameters. R23 and R8 identify the dish and medium bottle; the unspecified transfer volume remains null. \textbf{e}, Thirteen action interfaces and examples of reagent retrieval and liquid transfer. \textbf{f}, Action scores for 119 paired steps from 31 protocol blocks using Qwen 3.6 Flash and DeepSeek V4 Flash. Scores are normalized to applicable rubric maxima and averaged with equal step weights. In \textbf{c,f}, bars show means, dots individual scores and error bars 95\% confidence intervals from paired bootstrap resampling of tasks (\textbf{c}) or blocks (\textbf{f}).}
\end{minipage}}
\end{figure*}

\subsection{Intervention selection from biomedical evidence}
\label{sec:results_drug_repurposing}

At the intervention level, we asked whether biomedical evidence could guide drug-candidate selection for NMDA-induced retinal organoid injury. Starting from existing knowledge-graph links or predicted associations \cite{himmelstein2017hetionet,chandak2023primekg,gharizadeh2024hgtdr,zitnik2018decagon}, BioMARS used literature retrieval, refinement and ranking to assemble drug--target--mechanism--phenotype evidence chains, retaining supporting and contradictory evidence for experimental testing by researchers (Fig.~\ref{fig:drug}a). Of the six nominated candidates, memantine, agmatine, ifenprodil and milnacipran originated from existing one-hop links, whereas phenytoin and stiripentol originated from predicted associations.

The supporting evidence differed in directness. Memantine hydrochloride (Mem) had established NMDA-receptor pharmacology and prior evidence of protection against NMDA-induced retinal injury (Table~\ref{tab:mem_preserved_sources}); ifenprodil (Ife) and agmatine sulfate (Agm) had relatively direct NMDA-receptor pharmacological support \cite{williams2001ifenprodil,gibson2002agmatine}. Milnacipran hydrochloride (Mil) itself had pharmacological support for non-competitive NMDAR antagonism, with additional evidence of NMDA-channel blockade and neuroprotection from conformationally restricted analogues \cite{shuto2001evidence11325012,noguchi1999milnacipran}. Support for stiripentol (Sti) and phenytoin (Phe) came primarily from downstream pathways or other disease models \cite{bacq2024stiripentol,mantegazza2010phenytoin}. Representative evidence is shown in Fig.~\ref{fig:drug}c, and complete evidence profiles are provided in Table~\ref{tab:drug_evidence_portfolio}.

On differentiation day 14, retinal organoids were simultaneously exposed to 15~mM NMDA and the indicated compounds for 24~h (Fig.~\ref{fig:drug}b) \cite{hardingham2010nmda,papouin2012nmda}. NMDA exposure reduced area-normalized ATP signals compared with untreated controls. As a positive pharmacological control, 10~$\mu$M MK-801 consistently yielded higher area-normalized ATP signals than NMDA treatment alone across three independent experiments, attenuating the NMDA-associated reduction in this readout. Given the established neuroprotective effects of MK-801 against NMDA-induced retinal excitotoxicity \cite{eastlake2026muller}, this concordant pharmacological response supports the biological relevance of area-normalized ATP signals as a preliminary screening readout for NMDA-associated injury and potential protective effects (Fig.~\ref{fig:drug}e).

To evaluate the responses of the six candidates in the retinal organoid model, all compounds underwent preliminary ATP-based viability screening at concentrations of 1, 10 and 100~$\mu$M (Fig.~\ref{fig:drug}d). Memantine hydrochloride (Mem), agmatine sulfate (Agm) and phenytoin (Phe) showed higher mean area-normalized ATP signals than the corresponding NMDA reference groups at selected concentrations, whereas the remaining three candidates showed no consistent improvement. Based on these preliminary observations, Mem, Agm and Phe were selected for two additional independent experiments each, resulting in three experiments per compound to further evaluate their ATP-based viability responses under NMDA exposure.

Follow-up experiments with Mem, Agm and Phe were evaluated separately, and their concentration-resolved area-normalized ATP signals are summarized in Table~\ref{tab:drug_response_summary}. Together with the pharmacological response observed for the MK-801 positive control, the increases in area-normalized ATP signals observed for the three candidates at selected concentrations are consistent with potential attenuation of NMDA-induced injury and provide preliminary evidence supporting further evaluation of their protective effects (Fig.~\ref{fig:drug}e; Fig.~\ref{fig:drug_independent_experiments}).

%% file: sections/discussion.tex
\section{Discussion}
\label{sec:discussion}

BioMARS brings literature-grounded protocol formulation, feedback-driven robotic culture and intervention selection into a shared framework. Three separately evaluated studies address decisions through which biological experiments are formulated, adapted and tested. Building on robot scientists, self-driving laboratories and scientific agents \cite{king2004robotscientist,burger2020mobile,szymanski2023alab,boiko2023autonomous,swanson2025virtuallab,gottweis2026coscientist,ghareeb2026robin}, this work connects knowledge-guided condition selection with robotic culture, allowing measured biological feedback to inform subsequent experiments.

At the procedural level, BioMARS connects literature retrieval with explicit planning of experimental structure. Operations, recipes and timing requirements from different passages enter a common plan and remain linked through step identifiers in the final text. These links allow researchers to trace generated steps to the source literature while following the experimental stages and relationships between operations. Linking protocol steps to action objects and parameters extends this traceability to proposed laboratory operations.

The culture study jointly considered culture additives, initial cell number and resuspension procedures during condition selection. Literature knowledge informed the search, whereas measured morphological feedback guided subsequent combinations, allowing robotic culture conditions to change in response to experimental outcomes. The system thus selected combinations of formulations and procedures rather than optimizing each factor separately.

The intervention study translated pharmacological evidence from different sources into testable candidates for the organoid injury context. Candidates with direct NMDA pharmacology examined whether established effects could transfer to this system, whereas those supported by predicted associations or indirect evidence tested more extrapolative relationships. Directionally consistent increases in area-normalized ATP signals with memantine, agmatine sulfate and phenytoin during NMDA co-exposure provide exploratory support for further investigation of these candidates.

Cell-type-specific protection and mechanisms of action require further experimental validation.

Extending this framework to other biological systems requires matching literature evidence, experimental operations, phenotype readouts and safety constraints to the question being addressed. BioMARS provides a framework for this task-specific integration under human supervision.

\newgeometry{left=13.5mm,right=13.5mm,top=15mm,bottom=15mm,bindingoffset=0mm}
\begin{figure*}[p]
\centering
\makebox[\textwidth][c]{%
\begin{minipage}{183mm}
\includegraphics[width=\linewidth]{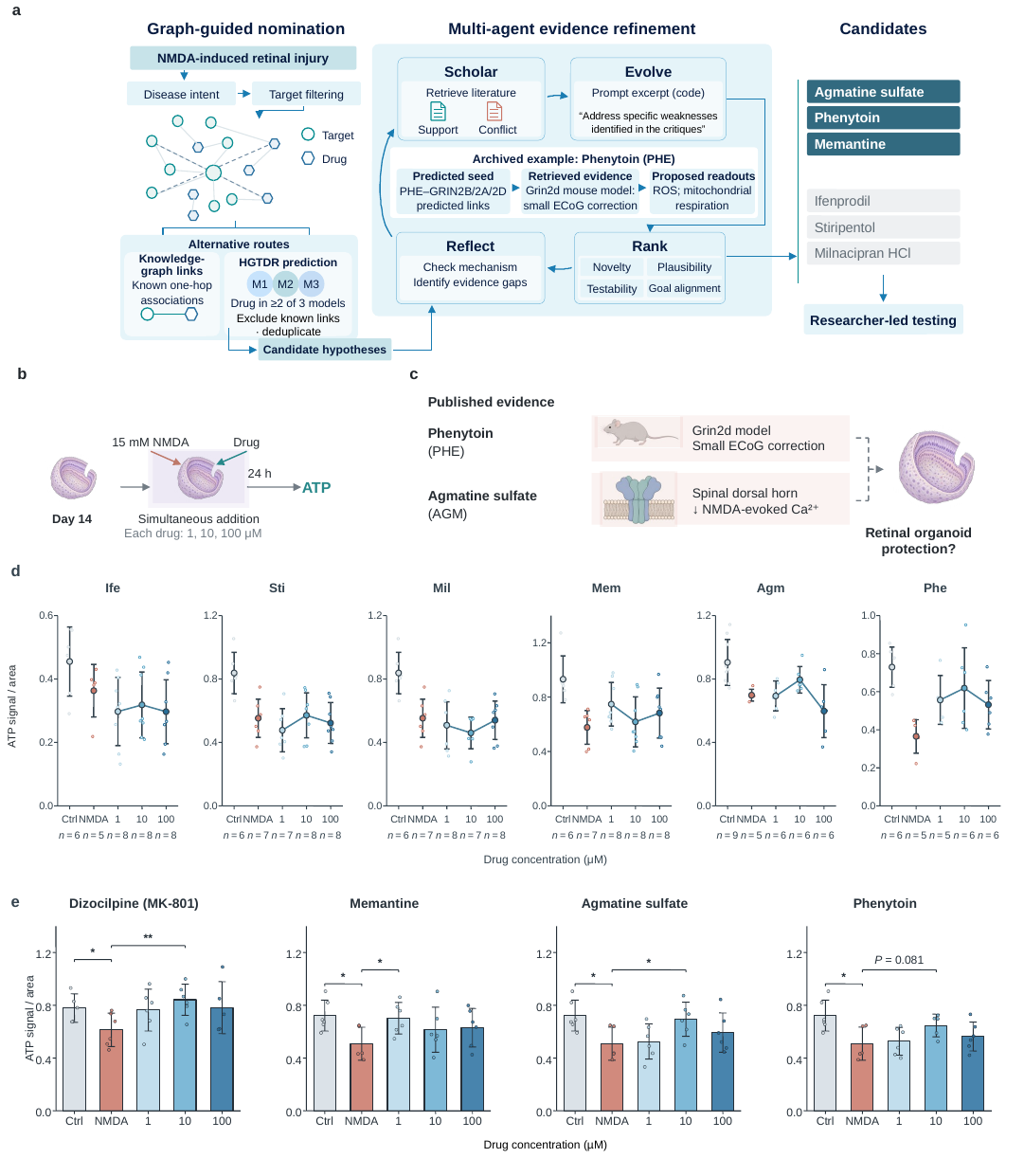}
\captionsetup{font=footnotesize,skip=1pt}
\caption{\textbf{Intervention nomination and retinal organoid injury testing.} \textbf{a}, Knowledge-guided candidate nomination and evidence refinement, illustrated with phenytoin. Dark labels identify candidates selected for follow-up. \textbf{b}, Day-14 organoids exposed simultaneously to 15~mM NMDA and drug (1, 10 or 100~$\mu$M) for 24~h. \textbf{c}, Published phenytoin and agmatine evidence from Grin2d mice and mouse spinal dorsal horn, respectively; dashed links indicate proposed effects in retinal organoids. \textbf{d}, Descriptive screening of six candidates. Sti and Mil share controls; Agm and Phe include cross-batch comparisons. Vertical-axis ranges differ. \textbf{e}, Follow-up tests of three candidates and MK-801, one experiment per plot. The Mem, Agm and Phe experiments shown share the same Ctrl and NMDA controls. In \textbf{d,e}, dots denote organoids; large circles (\textbf{d}) or bars (\textbf{e}) show means, with SD error bars. ATP signals are normalized to projected organoid area. Ctrl, normal medium; NMDA, NMDA alone. Group sizes are given in \textbf{d} and Table~\ref{tab:drug_response_summary}. Comparisons in \textbf{e} use two-sided Welch t-tests with unadjusted P values: $^{*}P<0.05$, $^{**}P<0.01$. Additional experiments are shown in Fig.~\ref{fig:drug_independent_experiments}.}
\label{fig:drug}
\end{minipage}}
\end{figure*}
\clearpage
\restoregeometry
\FloatBarrier

%% file: sections/methods.tex
\section{Methods}
\label{sec:methods}

\subsection{Study design and system architecture}
\label{sec:methods_system}

The three studies shared task orchestration, session state and linked records on an architecture for automated biological handling and culture \cite{dettinger2022open,kanda2022robotic,tristan2021robotic}. They were evaluated separately rather than as successive stages applied to the same samples.

Researchers defined scientific objectives, permitted procedures and safety constraints. The protocol-generation study evaluated text outputs intended for review before laboratory use. For culture optimization, researchers specified the factor space, experimental budget, phenotype scoring and aggregation rules, whereas the optimizers selected subsequent conditions. Preparation, handoffs and recovery remained under human oversight. Drug testing was performed by researchers. Decision and execution records for each task are described in Appendix section~\ref{sec:supp_loop_records}.

The physical platform integrates two six-axis and two four-axis robotic manipulators (Fig.~\ref{fig:system}b). Its stations provide reagent handling, single- and multichannel liquid transfer, container opening, incubation, cell counting and imaging. An additional mobile robot handles transport outside the platform. Station configurations, tip storage, waste collection and the dispensing workflow are shown in Figs.~\ref{fig:supp_hardware_workflow} and \ref{fig:supp_hardware_stations}.

\subsection{Robotic action interfaces}
\label{sec:methods_actions}

The action catalogue contains 13 action interfaces. These interfaces separate laboratory operations from lower-level manipulator movements, lid handling, pipetting, incubator handoffs and measurement routines \cite{steiner2019chemputer,mehr2020digitization}. Operation-specific arguments specify workstations, containers, reagents, volumes, wells or incubator positions; interface definitions are provided in Table~\ref{tab:action_interfaces}.

\subsection{Protocol generation and benchmarking}
\label{sec:methods_protocol}

Protocol generation comprised task definition, literature retrieval, structured planning, text generation and consistency checking, performed by Domain Pack, Research, Planning, Grounded Writer and Critic, respectively. Intermediate outputs were passed through shared task state. The first four stages used large language models (LLMs), whereas Critic applied deterministic rules.

Domain Pack extracted objectives and constraints from the task request and optional laboratory documents. Research retrieved and selected a primary source according to the starting material, requested operation and intended product, then extracted methods content while retaining the full article.

Planning segmented the article by headings and organized its content into five evidence views: operations, materials, recipes, timing and troubleshooting. Operations and temporal information were extracted with supporting quotations. Plans followed an experimental module--time node--step hierarchy, distinguishing stage progression, step durations and repetition schedules while retaining branch applicability.

After structural repair and task-scope selection, Grounded Writer generated step-level text, which was assembled by step identifier to preserve plan structure and citations. Critic checked step coverage, order, branches and numerical consistency. Full validation rules and intermediate products are described in Appendix section~\ref{sec:app_b1}.

The benchmark comprised 47 tasks evaluated with GPT-4o, Qwen 3.6 Flash, DeepSeek V4 Flash and Gemini 3 Flash Preview. Each task had a BioMARS and a FlashRAG \cite{jin2025flashrag} output for every model, yielding 376 protocols and 188 within-task, within-model pairs.

System-assisted scoring used a shared task-specific checklist based on the user query and source article; domain experts reviewed the assessments and confirmed the final scores. Five dimensions measured task-scope alignment (C), content accuracy (A), experimental logic (L), protocol structure (S) and procedural detail (D), with maxima of 30, 30, 20, 10 and 10 points, respectively (Table~\ref{tab:protocol_rubric}). Total scores were the sum of these dimensions.

BioMARS and FlashRAG scores were paired by task and model; relative gain was the mean score difference expressed as a percentage of the FlashRAG mean. Percentile 95\% confidence intervals were calculated from 20,000 paired-task bootstrap resamples within each model.

Confidence intervals describe task-sampling variability conditional on the ratings, not evaluator uncertainty. Dimension comparisons were descriptive. %

\subsection{Protocol-to-action mapping}
\label{sec:methods_protocol_action}

Action mapping used source steps and their Block context. Expansion specified experimental operations and associated handling, which were then mapped to predefined functions. Resource references linked samples and containers across calls, while arguments specified reagents, volumes, well positions and incubation conditions. The catalogue contained 13 functions covering material handling, liquid operations, culture and measurement (Table~\ref{tab:action_interfaces}).

Action evaluation used final protocol texts generated from the scientific literature, covering medium preparation, cell culture and passaging, differentiation, and organoid culture. Complete protocol sections were selected as Blocks with their context retained. Evaluation included 119 paired steps from 31 Blocks, selected for platform applicability and availability of generated outputs; source articles are listed in Source Data~1. System-assisted scoring compared source requirements with final actions across operation matching, known arguments, objects and handling, required order, and unknown-information handling, weighted at 30, 25, 25, 10 and 10 points, respectively. Each step was normalized to 100 using its applicable maximum, and the overall mean weighted steps equally. Scoring equations are provided in Appendix section~\ref{sec:supp_protocol_action}, and item-level assessments in Source Data~1.

Percentile 95\% confidence intervals for mean Action scores were calculated from 20,000 paired block-bootstrap resamples. Each resample drew 31 blocks with replacement, retaining all constituent steps and both model scores. Mean scores were recalculated across the pooled steps with equal step weights. The 2.5th and 97.5th percentiles defined the interval bounds.

\subsection{Retinal organoid culture}
\label{sec:methods_organoid}

All three culture-optimization rounds used OriCell 129 mouse embryonic stem cells (129 mESCs; MUAES-01001, Cyagen Biosciences). On day 0, cells were seeded in 96-well V-bottom ultra-low-attachment plates at the selected densities in retinal differentiation medium (RDM) containing the assigned concentrations of Y-27632, IGF-1 and progesterone (P4). Matrigel was added on day 1 to a final concentration of 2\% (v/v). Cultures were maintained without medium changes at 37~$^{\circ}$C with 5\% CO$_2$ until day 7. Bright-field images were acquired on day 4 for morphology-based scoring. Factor ranges, plate layouts, batches and aggregation rules are described in section~\ref{sec:methods_optimization}.

RDM was prepared by adding 5.1~ml each of 100$\times$ non-essential amino acid and 100$\times$ pyruvate stocks to 500~ml Glasgow minimum essential medium (GMEM). Additional supplements were 0.51~ml of 0.1~M 2-mercaptoethanol, 7.6~ml KnockOut Serum Replacement and 5.1~ml penicillin--streptomycin stock. Medium was sterilized through a 0.2-\ensuremath{\mu}m filter, stored at 4~$^{\circ}$C and used within one month.

Drug assays used separately cultured organoids with a fixed seeding density. Differentiation, maturation and treatment timing are described in section~\ref{sec:methods_injury_assays}. Organoids of similar size and morphology were selected before drug treatment. Optimization conditions were arranged by column, as specified in Table~\ref{tab:opt_plate_layout}. Plate position and processing order were retained for the optimization experiments to assess spatial and temporal effects.

\subsection{Image-derived phenotype scoring}
\label{sec:methods_phenotype_scoring}

Three biology experts reviewed 616 source images on the basis of morphological features. Of these, 472 were eligible for modelling: 376 failed images (label 0) and 96 successful images (label 2). We excluded 85 ambiguous images (label 1) and 59 images without valid information (label 3). Source images were partitioned before augmentation. The training set contained 352 failed and 72 successful images; validation and test sets each contained 12 images per class. We generated 280 augmented successful images within the training set only, producing 352 images per class and 752 records across all partitions.

A YOLO26m image classifier \cite{jocher2026yolo26} was trained on 640-pixel, tightly cropped regions of interest (ROIs). Training used AdamW with a cosine learning-rate schedule and early stopping. Inference used a medium ROI crop. Key settings are summarized in Table~\ref{tab:phenotype_configuration}; detailed parameters and model-selection records are provided in the accompanying reproducibility configuration file.

For continuous phenotype scoring, each image embedding $\mathbf{x}$ returned by the classifier was first L2-normalized, $\mathbf{z}=\mathbf{x}/\lVert\mathbf{x}\rVert_2$. The reference bank comprised all 472 eligible, unaugmented source images (96 successful and 376 failed). For class $c\in\{\mathrm{success},\mathrm{failure}\}$, Euclidean distances to its reference embeddings and their mean Gaussian radial basis function (RBF) density were
\begin{equation}
d_i^{(c)}=\lVert\mathbf{z}-\mathbf{z}_i^{(c)}\rVert_2, \qquad
D_c(\mathbf{z})=\frac{1}{N_c}\sum_{i=1}^{N_c}\exp\!\left(-\frac{(d_i^{(c)})^2}{2\sigma^2}\right).
\end{equation}
The score was then
\begin{equation}
S(\mathbf{z})=\frac{D_{\mathrm{success}}(\mathbf{z})}{D_{\mathrm{success}}(\mathbf{z})+D_{\mathrm{failure}}(\mathbf{z})}.
\end{equation}
The bandwidth was twice the median 15th-nearest-neighbour distance within the reference bank ($k=15$, bandwidth factor $=2.0$). The score measures morphological embedding similarity relative to this bank, not a calibrated probability of success or a direct measure of biological function. If both densities were numerically zero, the score was stored as NaN. Blank or invalid images likewise remained missing; only explicitly recorded experimental failures were assigned 0 during optimization aggregation.

\subsection{Retinal organoid culture optimization}
\label{sec:methods_optimization}

The prospective space contained six factors set on day 0. Chemical factors were Y-27632 (0--50~$\mu$M), IGF-1 (0--100~ng~ml$^{-1}$) and progesterone (P4; 0--10~$\mu$M). Physical factors were initial cell number (400--10,000 cells per well), resuspension cycles (0--10) and flow rate (40--200~$\mu$l~s$^{-1}$).

MASBO selected ten conditions per round over three rounds of robotic organoid experiments. After rounds 1 and 2, phenotype scores available at the time informed subsequent condition selection.

MASBO combined offline literature knowledge with online experimental feedback to select culture conditions. The offline stage selected literature passages using TF--IDF cosine similarity to derive factor-wise guidance and recommended regions for initial and subsequent proposals within the predefined bounds. Once experimental results were available, the LLM combined this guidance with tested conditions and their measured scores to propose combinations near high-scoring conditions and in knowledge-supported but sparsely sampled regions. Candidates were mapped to the executable experimental grid, checked against constraints and deduplicated before ranking. Factor ranges, step sizes, grid levels and integer constraints are provided in the Factor ranges sheet of Source Data 2.

A discriminative LLM surrogate adapted from LLAMBO \cite{liu2024llambo} repeatedly predicted candidate scores using measured history as context. A numerical routine calculated expected improvement relative to the best measured score from the prediction mean and standard deviation. The highest-ranking unselected candidate was accepted at each step until a batch of ten conditions was obtained. Prediction spread was used in the acquisition calculation, not treated as calibrated uncertainty; only measured outcomes were added to completed experimental history. The knowledge-update mechanism is described in Appendix section~\ref{sec:app_c2}.

Conventional Bayesian optimization (BO) provided a comparator, also selecting ten conditions per round over three rounds \cite{jones1998ego,shahriari2016bo,frazier2018bo,hase2018phoenics,shields2021reactionbo}. The two methods therefore tested 30 conditions each, giving 60 conditions in total. BO likewise used the available phenotype scores after rounds 1 and 2, without researchers manually choosing its subsequent proposals. Complete comparisons are reported in Table~\ref{tab:optimization_round_summary} and the accompanying sensitivity analyses. Retrospective comparisons of BO, LLAMBO and MASBO on Iris and iPSC datasets are reported separately (Appendix section~\ref{app:opt_experiment}) and were not pooled with the robotic organoid experiments.

Each condition was tested in three independent culture batches, with one plate per batch. Candidate scores used rows B--G and columns 2--11; control columns 1 and 12 and edge rows A and H were excluded (Table~\ref{tab:opt_plate_layout}). Masked invalid wells and blank measurements were excluded before aggregation. Three recorded condition-level failures (BO round 1 W09, MASBO round 1 W02 and BO round 3 W11) were assigned score 0. For these conditions, remaining well measurements were not averaged; original records were retained.

Valid well scores were first averaged for each condition within each plate. Pearson correlations were then calculated between the condition-mean vectors for all three plate pairs. The pair with the highest correlation was selected, and its corresponding condition means were averaged with equal weight to obtain the primary scores, without intermediate rounding. The third plate was retained for quality control. The concordance correlation coefficient (CCC) and mean absolute error described agreement but did not determine plate selection. Third-round sensitivity analyses used the mean and median across all three plates. Score distributions were summarized by condition.

Third-round analyses used image-derived well scores reported to two decimal places; cross-round scores are reported descriptively for this sequential experiment. Source Data 2 records plate-to-policy assignments; condition identifiers link well scores to six-factor settings. Reported scores, invalid-well masks and analysis values are provided separately in the source data.

Figure~\ref{fig:feedback}g shows one valid MASBO well per round to illustrate morphological variation, labelled with its individual well score. The images depict different organoids. Each original 1,920~$\times$~1,200-pixel image was cropped to the central 400~$\times$~400 pixels without intensity adjustment. Physical scale calibration was unavailable. Plate and well identifiers are provided in Appendix section~\ref{sec:app_c1}.

\subsection{Drug nomination and evidence review}
\label{sec:methods_drug}

A disease-intent agent retrieved pathogenesis evidence for retinal injury and mapped candidate genes and proteins to PrimeKG (Harvard Dataverse, dataset version 2.1; DOI: 10.7910/DVN/IXA7BM) \cite{chandak2023primekg}. A relation-filter agent retained central mapped targets. Drug candidates came from two routes: retrieval of existing one-hop drug--target links or prediction of unobserved associations by an HGTDR ensemble \cite{gharizadeh2024hgtdr}. Predicted candidates had to recur across model outputs, with known associations and duplicate pairs excluded. Prediction scores were normalized within each target and aggregated by drug using the mean and maximum normalized scores and target coverage. An LLM ranking agent ordered candidates by target relevance and biological plausibility.

Candidate hypotheses were refined through critique, literature retrieval, revision and ranking \cite{gottweis2026coscientist,internagent2025}, retaining supporting and contradictory evidence with its provenance. Ranking considered novelty, biological or clinical plausibility, experimental testability and alignment with the objective. Each hypothesis recorded drug--target--mechanism--phenotype links and evidence gaps. The computational workflow is detailed in Appendix~\ref{app:drug_repurposing}.

Evidence review distinguished direct drug--target pharmacology, analogue-based support and extrapolation from downstream pathways or other disease models. Graph provenance, model scores, evidence directness, novelty and observed efficacy were recorded separately. Evidence directness was not treated as a predictor of experimental response (Table~\ref{tab:drug_evidence_portfolio}).

\subsection{Retinal organoid injury assays}
\label{sec:methods_injury_assays}

Drug-screening experiments were conducted independently of robotic culture optimization using retinal organoids derived from OriCell™ 129 mouse embryonic stem cells (129 mESCs; Cyagen Biosciences, MUAES-01001). On day 0, 3,000 mESCs were seeded per well in 96-well V-bottom ultra-low-attachment plates containing 100~$\mu$l of retinal differentiation medium (RDM) and cultured at 37~$^{\circ}$C with 5\% CO$_2$. Matrigel was added on day 1 at a final concentration of 2\% (v/v). On day 7, organoids were transferred to 100-mm ultra-low-attachment dishes and maintained in retinal maturation medium (RMM).

On differentiation day 14, N-methyl-D-aspartic acid (NMDA; Sigma-Aldrich, M3262) was prepared in RMM and co-administered with the candidate compounds at a final concentration of 15~mM for 24~h. This concurrent-exposure design was informed by previous retinal organoid toxicity studies employing simultaneous exposure to toxic agents and pharmacological interventions \cite{rosarda2023imbalanced}. Six candidate compounds were evaluated at 1, 10 and 100~$\mu$M: ifenprodil tartrate (Ife; TargetMol, T1186), milnacipran hydrochloride (Mil; TargetMol, T1189), agmatine sulfate (Agm; TargetMol, T4808), stiripentol (Sti; TargetMol, T4498), phenytoin (Phe; TargetMol, T0939) and memantine hydrochloride (Mem; TargetMol, T1443). Experimental controls included untreated organoids, organoids exposed to NMDA alone and organoids co-treated with NMDA and dizocilpine (MK-801; MedChemExpress, HY-15084B) as a positive pharmacological control. MK-801 was likewise tested at 1, 10 and 100~$\mu$M.

After 24~h of treatment, organoids were individually transferred to white 96-well assay plates, with one organoid per well. Organoid viability was assessed using the CellTiter-Glo® 3D Cell Viability Assay (Promega, G9683) by measuring ATP-dependent luminescence as an indicator of metabolically active viable cells. The ATP-based assay was performed with reference to previously reported retinal organoid viability measurements \cite{mcclements2022tropism}, with reagent handling based on the manufacturer's protocol and an established three-dimensional organoid assay procedure \cite{promega2023tm412,baptista2023polymer}. CellTiter-Glo® 3D Reagent was added to each sample at a volume equal to that of the culture medium, followed by orbital mixing for 5~min to facilitate organoid lysis and incubation at room temperature for 25~min before luminescence measurement.

Bright-field images were acquired before the ATP assay to quantify each organoid's two-dimensional projected area in pixels (pixel$^2$). ATP-dependent luminescence was normalized to the corresponding area \cite{mcclements2022tropism,vergara2017three} as follows:

\[
\text{Area-normalized ATP signal}
=
\frac{\mathrm{luminescence}}
{\mathrm{projected\ organoid\ area}}.
\]

All six candidate compounds underwent preliminary ATP-based viability screening. Based on the ATP responses observed in the preliminary screen, agmatine sulfate, phenytoin and memantine hydrochloride were selected for two additional independent experiments each, resulting in three experiments per selected compound. The remaining three candidates were evaluated only in the preliminary screen. MK-801 was assessed separately in three experiments as the positive pharmacological control.

Candidate drug stocks were prepared in dimethyl sulfoxide (DMSO) or sterile water; MK-801 was prepared as a 50~mM stock in DMSO. Stocks were serially diluted in culture medium to the treatment concentrations without additional solvent, and solvent-matched vehicle controls were not included. Stock formulations and final DMSO fractions calculated from dilution factors are provided in Table~\ref{tab:drug_stock_vehicles}.

%% file: appendices/appendices.tex
\clearpage
\begin{appendices}
\makeatletter
\setlength{\@fptop}{0pt}
\setlength{\@fpsep}{14pt}
\setlength{\@fpbot}{0pt plus 1fil}
\setlength{\@dblfptop}{0pt}
\setlength{\@dblfpsep}{14pt}
\setlength{\@dblfpbot}{0pt plus 1fil}
\makeatother
\renewcommand{\topfraction}{0.95}
\renewcommand{\bottomfraction}{0.90}
\renewcommand{\textfraction}{0.05}
\renewcommand{\floatpagefraction}{0.80}
\renewcommand{\dbltopfraction}{0.95}
\renewcommand{\dblfloatpagefraction}{0.80}
\setlength{\textfloatsep}{14pt plus 2pt minus 2pt}
\setlength{\floatsep}{12pt plus 2pt minus 2pt}
\setlength{\dbltextfloatsep}{14pt plus 2pt minus 2pt}

\section*{Appendix guide}
\renewcommand{\theHfigure}{appendix.\thesection.\arabic{figure}}
\renewcommand{\theHtable}{appendix.\thesection.\arabic{table}}

Appendices A--D provide the evaluations, data-processing details and implementation records supporting the three tasks. Protocol assessment, culture-score analysis and drug-candidate evidence are reported separately, with experimental scope specified in the relevant sections.

\begingroup
\small
\renewcommand{\arraystretch}{1.12}
\noindent
\begin{tabularx}{\textwidth}{@{}lXr@{}}
\textbf{Section} & \textbf{Contents} & \textbf{Page} \\
\midrule
\textbf{A} & \textbf{BioMARS platform and laboratory interfaces} & \pageref{app:platform} \\
A.1 & Laboratory hardware and liquid handling & \pageref{sec:app_a2} \\
A.2 & Laboratory action interfaces & \pageref{sec:app_a3} \\
A.3 & Decision and execution records & \pageref{sec:app_a4} \\
\textbf{B} & \textbf{Protocol generation and action mapping} & \pageref{app:protocol_generation} \\
B.1 & Literature-grounded protocol generation & \pageref{sec:app_b1} \\
B.2 & Protocol quality assessment & \pageref{sec:app_b2} \\
B.3 & Protocol-to-action mapping & \pageref{sec:supp_protocol_action} \\
\textbf{C} & \textbf{Phenotype-guided culture optimization} & \pageref{app:culture_optimization} \\
C.1 & Image-derived phenotype scoring & \pageref{sec:app_c1} \\
C.2 & MASBO implementation & \pageref{sec:app_c2} \\
C.3 & Three-round culture optimization & \pageref{sec:app_c3} \\
C.4 & Plate scores and quality control & \pageref{sec:app_c4} \\
C.5 & Factor space and sensitivity analyses & \pageref{sec:app_c5} \\
C.6 & Retrospective optimization benchmarks & \pageref{sec:app_c7} \\
\textbf{D} & \textbf{Drug repurposing and experimental evaluation} & \pageref{app:drug_repurposing} \\
D.1 & Graph-guided candidate nomination & \pageref{sec:app_d1} \\
D.2 & Candidate evidence and evidence-chain assessment & \pageref{sec:app_d2} \\
D.3 & Drug stocks and vehicles & \pageref{sec:app_d4} \\
D.4 & Preliminary screening and follow-up & \pageref{sec:app_d5} \\
\end{tabularx}
\endgroup
\clearpage

\input{appendices/appendix_a_platform}

\FloatBarrier

\input{appendices/appendix_b_protocol_generation}

\FloatBarrier

\input{appendices/appendix_c_culture_optimization}
\FloatBarrier

\input{appendices/appendix_d_drug_repurposing}

\FloatBarrier
\clearpage

\end{appendices}

%% file: appendices/appendix_a_platform.tex
\section{BioMARS platform and laboratory interfaces}
\label{app:platform}
\label{app:extended_data}
\label{app:computational_implementation}
\setcounter{figure}{0}
\setcounter{table}{0}

\FloatBarrier
\subsection{Laboratory hardware and liquid handling}
\label{sec:app_a2}

\begin{figure}[!htbp]
\centering
\includegraphics[width=0.90\textwidth,height=0.88\textheight,keepaspectratio]{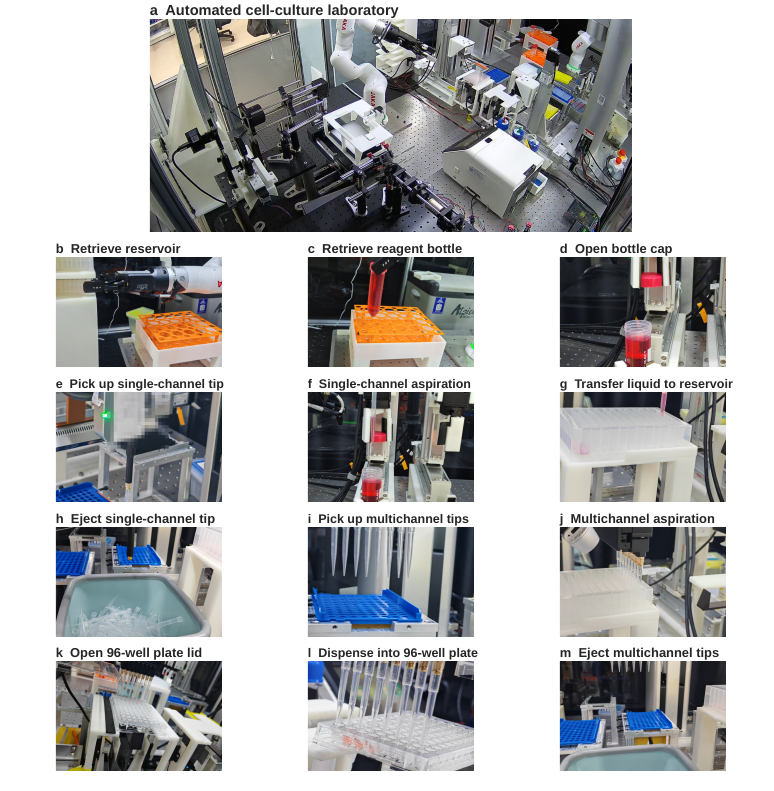}
\caption{\textbf{Reagent dispensing in the automated laboratory.} \textbf{a}, Automated cell-culture laboratory. \textbf{b--d}, Reservoir and reagent-bottle retrieval, followed by cap opening. \textbf{e--h}, Single-channel tip pickup, aspiration, transfer to the reservoir and tip ejection. \textbf{i--m}, Multichannel tip pickup, aspiration, 96-well-plate lid opening, dispensing and tip ejection.}
\label{fig:supp_hardware_workflow}
\end{figure}

\begin{figure}[!htbp]
\centering
\includegraphics[width=\textwidth,height=0.88\textheight,keepaspectratio]{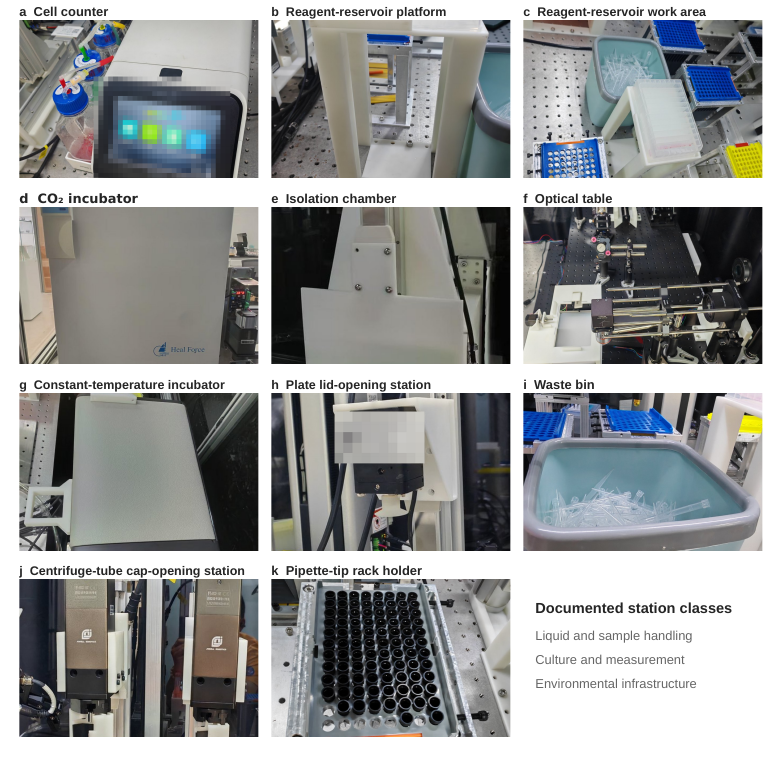}
\caption{\textbf{Laboratory stations and infrastructure.} \textbf{a--k}, Cell counter, reagent-reservoir platform, reagent-reservoir work area, CO$_2$ incubator, isolation chamber, optical table, constant-temperature incubator, plate lid-opening station, waste bin, centrifuge-tube cap-opening station and pipette-tip rack holder, respectively.}
\label{fig:supp_hardware_stations}
\end{figure}
\FloatBarrier

\FloatBarrier
\subsection{Laboratory action interfaces}
\label{sec:app_a3}
\begingroup
\label{sec:supp_action_interfaces}

The current action catalogue defines 13 action interfaces for material handling, liquid operations, culture and measurement. Table~\ref{tab:action_interfaces} lists the functions, their main arguments and supported operations.

\begingroup
\footnotesize
\setlength{\tabcolsep}{4pt}
\renewcommand{\arraystretch}{1.12}
\begin{longtable}{@{}>{\raggedright\arraybackslash}p{0.28\textwidth}>{\raggedright\arraybackslash}p{0.30\textwidth}>{\raggedright\arraybackslash}p{\dimexpr0.42\textwidth-16pt\relax}@{}}
\caption{\textbf{Action catalogue used for protocol mapping.} Thirteen functions covering material handling, liquid operations, culture and measurement. Entries summarize function arguments and operation semantics.}\label{tab:action_interfaces}\label{tab:semantic_catalog}\\
\toprule
Function & Main arguments & Operation scope \\
\midrule
\endfirsthead
\multicolumn{3}{l}{Table~\thetable\ (continued)}\\
\toprule
Function & Main arguments & Operation scope \\
\midrule
\endhead
\bottomrule
\endfoot
Get container\newline \texttt{get\_\allowbreak{}new\_\allowbreak{}container\_\allowbreak{}from\_\allowbreak{}storage} & Storage location; target workstation; container type & Acquire a new receiving vessel. Acquisition does not fill, coat or load it. \\
\addlinespace[3pt]
Discard container\newline \texttt{discard\_\allowbreak{}container} & Source workstation; container type & Dispose of a vessel designated as waste; distinct from returning a vessel to storage. \\
\addlinespace[3pt]
Get reagent\newline \texttt{get\_\allowbreak{}reagent} & Reagent identity; container type; target workstation & Retrieve an existing reagent or sample in its container. \\
\addlinespace[3pt]
Return reagent\newline \texttt{return\_\allowbreak{}reagent} & Reagent identity; container type; workstation & Return the same reagent or sample container to storage after use. \\
\addlinespace[3pt]
Transfer liquid\newline \texttt{transfer\_\allowbreak{}liquid\_\allowbreak{}between\_\allowbreak{}workstations} & Source and target workstation/types; volume; addressed wells; optional mixing & Transfer liquid or suspension to the specified receiver. Plate addressing follows the supported column or well convention. \\
\addlinespace[3pt]
Resuspend cells\newline \texttt{resuspend\_\allowbreak{}cells\_\allowbreak{}in\_\allowbreak{}tube} & Tube workstation and type; medium and volume fields & Mix cells already in a supported tube. This call does not itself add medium or adjust concentration. \\
\addlinespace[3pt]
Prepare mixed solution\newline \texttt{prepare\_\allowbreak{}mixed\_\allowbreak{}solution} & Calculation mode; receiving container; total volume; recipe; diluent & Prepare a multicomponent mixture. Component handling is internal; the receiver is acquired separately and retains the product. \\
\addlinespace[3pt]
Dilute solution\newline \texttt{dilute\_\allowbreak{}solution} & Calculation mode; receiving container; total volume; stock/diluent recipe & Dilute an existing solution or cell suspension. Required unknown quantities remain pending. \\
\addlinespace[3pt]
Remove liquid\newline \texttt{remove\_\allowbreak{}liquid} & Source workstation/type; removal mode; volume or residual volume; addressed well & Remove liquid as waste. Default removal is all liquid when neither a removal nor residual amount is specified; replacement liquid is a separate operation. \\
\addlinespace[3pt]
Retrieve from incubator\newline \texttt{retrieve\_\allowbreak{}container\_\allowbreak{}from\_\allowbreak{}incubator} & Incubator position; target workstation; container type & Retrieve the existing culture vessel, retaining its sample identity. \\
\addlinespace[3pt]
Incubate container\newline \texttt{incubate\_\allowbreak{}container} & Source workstation; container type; incubator position; duration & Transfer a supported culture vessel into incubation. Upper-level scheduling is separate. \\
\addlinespace[3pt]
Count cells\newline \texttt{count\_\allowbreak{}cells} & Sample workstation; container type & Count cells in an available suspension; collecting the main sample and diluting it are separate operations. \\
\addlinespace[3pt]
Image plate\newline \texttt{image\_\allowbreak{}plate} & Sample workstation; optional plate name and day label & Acquire bright-field images. \\
\addlinespace[3pt]
\end{longtable}
\endgroup

Persistent resource identifiers link containers across operations. Models specify unnumbered workstation types; platform dispatch assigns physical positions. The addressing convention uses columns for 96-well plates and individual wells for six-well plates.

\endgroup

\FloatBarrier
\subsection{Decision and execution records}
\label{sec:app_a4}
\label{sec:supp_loop_records}

The protocol-generation workflow includes human review before laboratory use.

Separate provenance chains link source evidence to generated protocols, culture conditions to measurements, and nominated drugs to experimental outcomes. Detailed record fields are listed in the accompanying reproducibility configuration file.

%% file: appendices/appendix_b_protocol_generation.tex
\section{Protocol generation and action mapping}
\label{app:protocol_generation}
\setcounter{figure}{0}
\setcounter{table}{0}

\FloatBarrier
\subsection{Literature-grounded protocol generation}
\label{sec:app_b1}
The request retains the original query, experimental scope and constraints such as species and cell line. Domain Pack assembles laboratory documents, existing procedures and safety requirements, distinguishing known from unconfirmed information. Research searches Semantic Scholar, PubMed, official technical resources and protocols.io, deduplicates candidates and checks their relevance and procedural applicability to select a primary methods source. Full-text ingestion verifies document identity, extracts methods content and retains the full text, source identifiers and links.

Planning segments the source by heading structure, retains segment identifiers and organizes evidence into five views: operations, materials, recipes, timing and troubleshooting. Extracted operations and temporal information carry supporting quotations, whose occurrence in the input is checked programmatically. Experimental progression, individual step durations and repetition schedules are represented separately.

These evidence views feed a plan of modules, time nodes and ordered steps. Branch identifiers distinguish alternative experimental routes, and each step retains its source sentences and identifiers. The plan covers core operations, necessary prerequisites and requested readouts, places reagent preparation before use and retains operational constraints as notes.

Grounded Writer drafts individual steps and global explanatory sections from the plan and linked sources. Programmatic assembly organizes the text, headings, numbering and permitted citations by step identifier. Changes to values, units, entities or meaning are recorded separately from stylistic edits and checked against source evidence.

Critic compares the plan and generated text using deterministic checks for missing, repeated, unknown or reordered steps and for consistency of modules, time nodes, branches, sources, numerical facts, citations and formatting. A structured report records check status, issues and warnings. A shared orchestrator schedules Domain Pack, Research, Planning, Grounded Writer and Critic and retains session state; these roles can use the same underlying model.

\FloatBarrier
\subsection{Protocol quality assessment}
\label{sec:app_b2}
\label{sec:app_b3}
\label{sec:app_b4}

The benchmark comprises 47 tasks evaluated with four model backends under BioMARS and FlashRAG, yielding eight groups and 376 outputs. All outputs for a task share a source-based assessment checklist. Domain experts reviewed system-assisted assessments and confirmed the final scores. The total is the sum of the five dimension scores, with a maximum of 100 points (Table~\ref{tab:protocol_rubric}).

C sums the fixed checklist weights of required functions that are developed into procedures, with a two-point deduction for each explicitly unrelated experimental function developed into a procedure, to a minimum of zero. Related supplementary content incurs no scope deduction. A and L subtract independent defects from 30 and 20 points, respectively, to a minimum of zero; deduction levels are 1, 2 or 4 points for A and 1, 2 or 3 points for L. Each defect is assigned to one dimension.

S sums five document-level structure criteria, each scored from 0 to 2 in half-point increments. D is ten times the unweighted mean of parameter-retention fractions across eligible checklist groups. Item-level evidence, assessment eligibility and the full calculation rules accompany the source data.

\begin{table}[htbp]
\centering
\caption{\textbf{Protocol-assessment rubric and source-data codes.} The five criteria sum to 100 points. Each task uses a common source-based checklist across all models and pipelines. Domain experts reviewed system-assisted assessments and confirmed final scores.}
\label{tab:protocol_rubric}
\small
\begin{tabular}{p{0.06\textwidth}p{0.12\textwidth}p{0.69\textwidth}}
\toprule
Code & Maximum & Assessment criterion \\
\midrule
C & 30 & Task-scope alignment: coverage of required functions, with deductions for explicitly unrelated experimental functions. \\
A & 30 & Content accuracy: fidelity of described operations and key conditions to the source. \\
L & 20 & Experimental logic: consistency of samples, branches, ordering, repetitions and input--output relationships. \\
S & 10 & Protocol structure: hierarchy, step localization, reading order, information accessibility and concision. \\
D & 10 & Procedural detail: retention of applicable parameters in the operations described. \\
\bottomrule
\end{tabular}
\end{table}

Scores are paired by task and model, giving 188 comparisons. Absolute gain is the mean paired BioMARS-minus-FlashRAG difference; relative gain expresses this difference as a percentage of the FlashRAG mean. Within each model, 20,000 bootstrap resamples of the 47 task pairs yield percentile 95\% confidence intervals. Mean scores, absolute gains and relative gains use the same resamples.

\begin{table}[htbp]
\centering
\caption{\textbf{Paired protocol-assessment gains.} BioMARS and FlashRAG were evaluated on the same 47 tasks for each model. Scores range from 0 to 100 under the five-dimensional rubric, with system-assisted assessments reviewed and confirmed by domain experts. Absolute gains are paired mean differences; relative gains express the difference as a percentage of the FlashRAG mean. Parentheses show percentile 95\% confidence intervals from 20,000 paired-task bootstrap resamples. Counts are improved/tied/declined tasks.}
\label{tab:protocol_gains}
\small
\setlength{\tabcolsep}{3pt}
\begin{tabularx}{\textwidth}{@{}>{\raggedright\arraybackslash}Xrrllr@{}}
\toprule
Model & FlashRAG & BioMARS & Gain, points (95\% CI) & Gain, \% (95\% CI) & Counts \\
\midrule
GPT-4o & 73.80 & 86.71 & 12.91 (9.23, 16.53) & 17.5 (12.2, 23.1) & 40/1/6 \\
Qwen 3.6 Flash & 78.35 & 94.69 & 16.34 (13.44, 19.33) & 20.9 (16.6, 25.6) & 44/0/3 \\
DeepSeek V4 Flash & 81.98 & 93.89 & 11.92 (8.72, 15.28) & 14.5 (10.3, 19.2) & 40/1/6 \\
Gemini 3 Flash Preview & 76.51 & 90.11 & 13.60 (10.90, 16.40) & 17.8 (13.8, 22.1) & 42/1/4 \\
\bottomrule
\end{tabularx}
\end{table}

\begin{figure}[htbp]
\centering
\includegraphics[width=\textwidth]{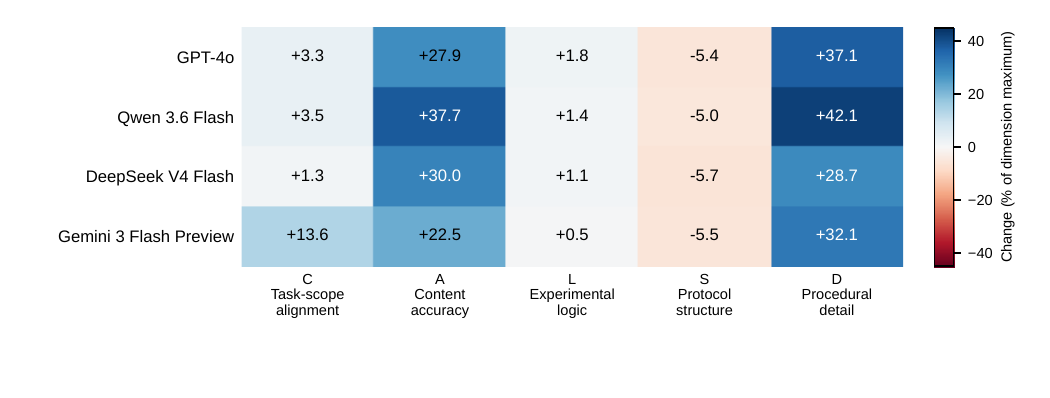}
\caption{\textbf{Changes in protocol-assessment dimensions.} Cells show mean paired BioMARS-minus-FlashRAG score differences, expressed as percentages of dimension maxima ($n=47$ tasks per model). Blue and red indicate positive and negative differences, respectively. This normalization differs from the relative mean gains in Fig.~\ref{fig:protocol}c.}
\label{fig:protocol_dimension_changes}
\end{figure}

Source Data 1 contains the 376 protocol ratings, 188 paired comparisons and 20 model-by-dimension comparisons. Each dimension row reports both pipelines' mean scores and their difference across 47 tasks.

\input{appendices/appendix_b_action_mapping}

%% file: appendices/appendix_b_action_mapping.tex
\subsection{Protocol-to-action mapping}
\label{sec:supp_protocol_action}

\subsubsection{Evaluation set}
The source collection comprised 206 steps from 38 protocol Blocks, with the surrounding context supplied for each step. Selection considered platform applicability, availability of generated outputs and applicability of the scoring rubric, yielding 119 paired steps from 31 Blocks. The Action evaluation set was selected and organized separately, with some source protocols overlapping with the 47-task protocol-generation benchmark.

\subsubsection{Scoring}
System-assisted assessment used the source step and its Block context as the reference for evaluating final actions. The five dimensions were operation matching (A, 30 points), known arguments (B, 25), objects and handling (C, 25), required order (D, 10), and unknown-information handling (E, 10).

For operation matching, true positives (TP) were required operations represented by appropriate functions, false negatives (FN) were missing operations, and false positives (FP) were unjustified additional core operations. The score was
\begin{equation}
A_s=30\,\frac{2\mathrm{TP}_s}{2\mathrm{TP}_s+\mathrm{FN}_s+\mathrm{FP}_s}.
\end{equation}
For each applicable dimension $d\in\{B,C,D,E\}$, the score was $P_{sd}=w_d p_{sd}/n_{sd}$, where $w_d$ was the dimension weight, $p_{sd}$ the number of passed checks and $n_{sd}$ the number of assessed pass/fail checks. The normalized step score was
\begin{equation}
S_s=100\,\frac{\sum_{d\in\mathcal{D}_s}P_{sd}}{\sum_{d\in\mathcal{D}_s}w_d},
\end{equation}
where $\mathcal{D}_s$ denotes the applicable dimensions. Overall scores were the unweighted means of the 119 step scores. The complete rubric and item-level judgments are supplied with Source Data~1.

\subsubsection{Dimension scores}
Table~\ref{tab:action_review} reports the overall and dimension scores. Operation matching scored higher than known arguments and object handling for both backends. In the medium-change example, both outputs included removal, transfer and incubation, but DeepSeek specified MEF medium instead of LIF + 2i medium. This difference was scored under known arguments.

\begin{table}[htbp]
\centering\small
\caption{\textbf{Step-level Action assessment.} Dimension scores are unweighted means on a 0--100 scale across applicable steps; $n$ is the number of scored steps for that dimension. The total is the mean normalized step score across all 119 steps; dimension means use their respective applicable subsets.}
\label{tab:action_review}
\begin{tabular*}{\textwidth}{@{\extracolsep{\fill}}lrrrrr}
\toprule
Dimension & Weight & Qwen $n$ & Qwen score & DeepSeek $n$ & DeepSeek score \\
\midrule
Operation matching & 30 & 119 & 94.29 & 119 & 93.20 \\
Known arguments & 25 & 66 & 83.33 & 67 & 71.64 \\
Objects and handling & 25 & 111 & 76.58 & 112 & 80.36 \\
Required order & 10 & 30 & 96.67 & 32 & 93.75 \\
Unknown information & 10 & 47 & 97.87 & 47 & 95.74 \\
\midrule
Total & 100 & 119 & 85.56 & 119 & 83.68 \\
\bottomrule
\end{tabular*}
\end{table}

%% file: appendices/appendix_c_culture_optimization.tex
\section{Phenotype-guided culture optimization}
\label{app:culture_optimization}
\setcounter{figure}{0}
\setcounter{table}{0}

\FloatBarrier
\subsection{Image-derived phenotype scoring}
\label{sec:app_c1}

\begin{center}
\begin{minipage}{\textwidth}
\captionsetup{type=table}
\begin{threeparttable}
\tablebodyfont
\centering
\caption{\textbf{Composition of the phenotype dataset.} Labels 0 and 2 were eligible for modelling; labels 1 and 3 were excluded. Counts distinguish unique source images from augmented training records. Augmentation was restricted to the successful class in the training partition.}
\label{tab:phenotype_dataset}
\begin{tabular}{p{0.30\textwidth}p{0.16\textwidth}p{0.45\textwidth}}
\toprule
Record & Number of images & Definition or use \\
\midrule
All human-reviewed source images & 616 & Images assigned one of four review labels \\
Failed, label 0 & 376 & Eligible reference and modelling class \\
Successful, label 2 & 96 & Eligible reference and modelling class \\
Ambiguous/uncertain, label 1 & 85 & Excluded from modelling \\
No valid information, label 3 & 59 & Excluded from modelling \\
Training, source images & 424 & 352 failed and 72 successful \\
Training, offline augmentation & 280 & Additional successful-class images only \\
Training, final & 704 & 352 failed and 352 successful \\
Validation & 24 & 12 failed and 12 successful; no augmentation \\
Nominal test & 24 & 12 failed and 12 successful; no augmentation \\
Final manifest & 752 & Training, validation and nominal test records \\
\bottomrule
\end{tabular}
\end{threeparttable}
\end{minipage}
\end{center}

\begin{center}
\begin{minipage}{\textwidth}
\captionsetup{type=table}
\begin{threeparttable}
\tablebodyfont
\centering
\caption{\textbf{Key settings for classifier training and RBF scoring.} RBF denotes radial basis function. Scores range from 0 to 1, with higher values indicating greater relative similarity to successful references. Detailed training and augmentation parameters are provided in the accompanying reproducibility configuration file.}
\label{tab:phenotype_configuration}
\begin{tabular}{p{0.31\textwidth}p{0.60\textwidth}}
\toprule
Component & Key settings \\
\midrule
Classifier initialization & YOLO26m; classification task; input size 640 pixels \\
Optimization & AdamW; initial learning rate 0.0003 \\
Training schedule & Cosine learning-rate schedule with early stopping \\
Training-time augmentation & Rotation, translation, scaling, flips and brightness adjustment \\
Offline augmentation & Geometric and intensity transformations, contrast enhancement, noise and sharpening; applied only to successful-class training images \\
RBF reference bank & 472 unaugmented source images: 96 successful and 376 failed \\
Phenotype score & Success-reference density divided by the sum of success- and failure-reference densities \\
Missingness & Zero density denominator stored as NaN; blank/no-valid-image cases are not assigned 0 \\
\bottomrule
\end{tabular}
\end{threeparttable}
\end{minipage}
\end{center}

The images in Fig.~\ref{fig:feedback}g came from unmasked MASBO wells: W09/G9 on MASBO--Round1--Plate2, W05/F5 on MASBO--Round2--Plate1, and W07/B7 on MASBO--Round3--Plate2. Labels report individual well image scores rather than aggregated condition scores. Image selection and cropping are described in section~\ref{sec:methods_optimization}.

\FloatBarrier
\subsection{MASBO implementation}
\label{sec:app_c2}
Let the completed wet-lab data after round $t$ be
\[
\mathcal{D}_t=\{(\mathbf{x}_i,y_i)\}_{i=1}^{n_t},
\]
where $\mathbf{x}_i$ contains the six factor settings and $y_i$ is the accepted image-derived phenotype score. The implementation receives three further inputs: the experiment description and endpoint, a factor schema defining names, units, bounds and executable value grids, and factor-wise knowledge generated before proposal selection.

The offline knowledge component constructs background, factor-specific and interaction queries from the experiment and factor schema. Retrieved abstracts or full text are divided into passages and indexed using TF--IDF vectors. For a query vector $\mathbf{v}_q$ and passage vector $\mathbf{v}_d$, retrieval uses cosine similarity,
\[
\operatorname{sim}(q,d)=
\frac{\mathbf{v}_q\cdot\mathbf{v}_d}
{\lVert\mathbf{v}_q\rVert\lVert\mathbf{v}_d\rVert},
\]
and the highest-scoring passages inform recommended regions and their supporting evidence for each factor. These records and the measured history $\mathcal{D}_t$ are supplied to the candidate-generation model without changing the experimental bounds.

For every plate position to be proposed, the acquisition model receives the observed configurations and scores, task context, hard factor constraints and factor knowledge. Each request is divided approximately 2:1 between exploitation around high-performing observations and exploration of less sampled regions. Raw candidates are then projected to the executable grid. For factor $k$, with lower bound $l_k$ and step $s_k$, the projection is
\[
Q_k(x)=l_k+\operatorname{round}\!\left(\frac{x-l_k}{s_k}\right)s_k,
\]
followed by clipping to the configured bounds. Candidate signatures are formed from all six quantized values to remove duplicate combinations within the batch after quantization and prevent reselection of combinations already accepted for the same plate.

The remaining pool is evaluated by a discriminative LLM surrogate. Repeated model score predictions for candidate $\mathbf{x}$ yield an empirical mean $\mu(\mathbf{x})$ and standard deviation $\sigma(\mathbf{x})$. For maximization with best observed value $y^+$, selection uses expected improvement,
\[
\operatorname{EI}(\mathbf{x})=
(\mu(\mathbf{x})-y^+)\Phi(z)+\sigma(\mathbf{x})\phi(z),
\qquad
z=\frac{\mu(\mathbf{x})-y^+}{\sigma(\mathbf{x})}.
\]
The unique candidate with the highest EI is accepted and blocked against reselection. Generation and selection continue until all plate positions are filled. Predicted scores are retained as computational records and are not treated as measured outcomes. The proposal is converted into six-factor settings for robotic plate preparation. Tested conditions and their scores are provided in the source data; model configurations and code access are addressed in Methods and Code availability.

The implemented knowledge-update mechanism compares a current result with the best historical observation. The Reflector proposes revisions to factor guidance specifying the parameter, direction, absolute range and supporting evidence; the Curator consolidates these suggestions. The dashed path in Fig.~\ref{fig:feedback}a represents the knowledge-update mechanism.

\FloatBarrier
\subsection{Three-round culture optimization}
\label{sec:app_c3}
Table~\ref{tab:optimization_round_summary} reports complete round-wise outcomes for both policies; plate-quality and sensitivity analyses follow below.

\input{tables/tab_opt_round_summary}

\FloatBarrier
\subsection{Plate scores and quality control}
\label{sec:app_c4}
\begin{figure}[htbp]
\centering
\includegraphics[width=0.86\textwidth]{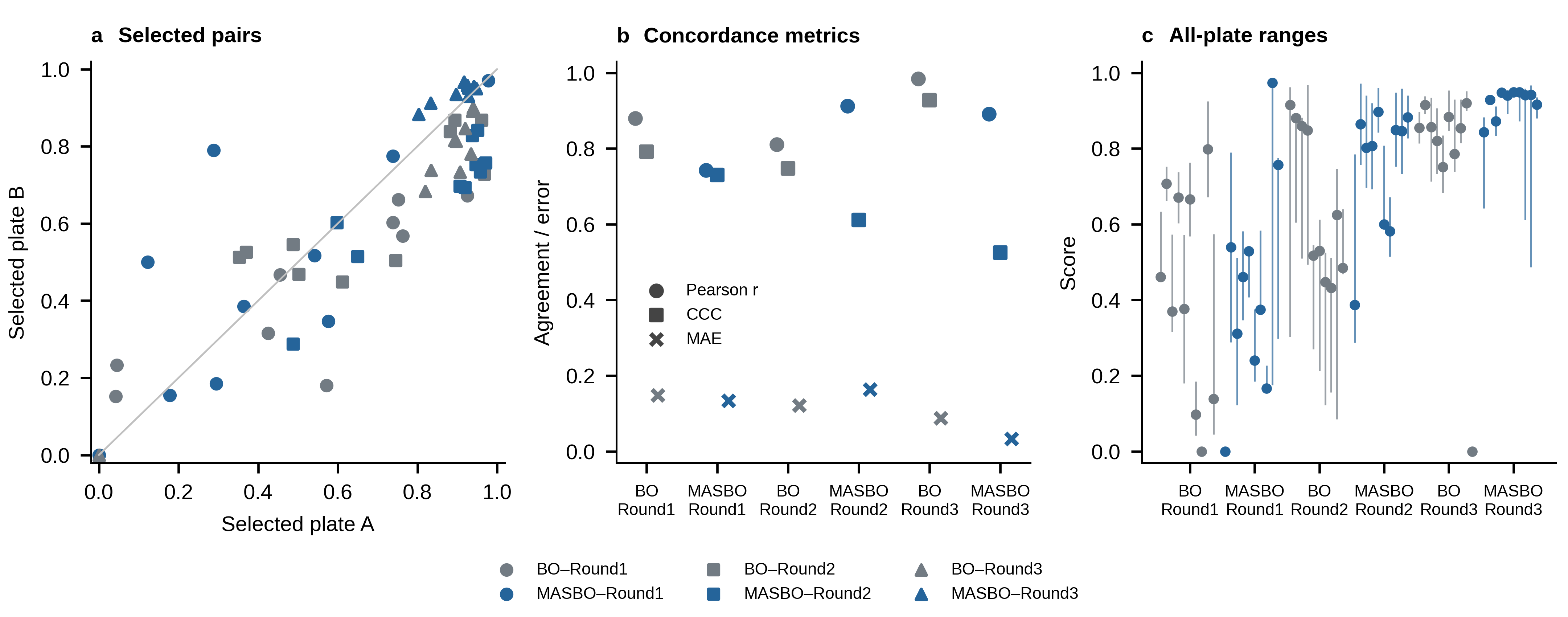}
\caption{\textbf{Plate-level concordance and variation.} \textbf{a}, Scores from the most highly correlated plate pair for each method and round ($n=10$ conditions); the diagonal indicates equality. \textbf{b}, Pearson correlation, concordance correlation coefficient (CCC) and mean absolute error (MAE) for these pairs. \textbf{c}, Three-plate score ranges, with points indicating selected-pair means. Grey, BO; blue, MASBO.}
\label{fig:plate_concordance}
\end{figure}

Plate-pair concordance and third-plate deviations are shown in Fig.~\ref{fig:plate_concordance}; all pairwise values appear in Source Data 2. The selected pairs were BO--Round3--Plate2 and BO--Round3--Plate3 ($r=0.985$), and MASBO--Round3--Plate2 and MASBO--Round3--Plate3 ($r=0.891$). Table~\ref{tab:optimization_round_summary} reports the selected pairs across all three rounds; plate-level records are provided in Source Data 2.

\begin{figure}[!htbp]
\centering
\includegraphics[width=\textwidth]{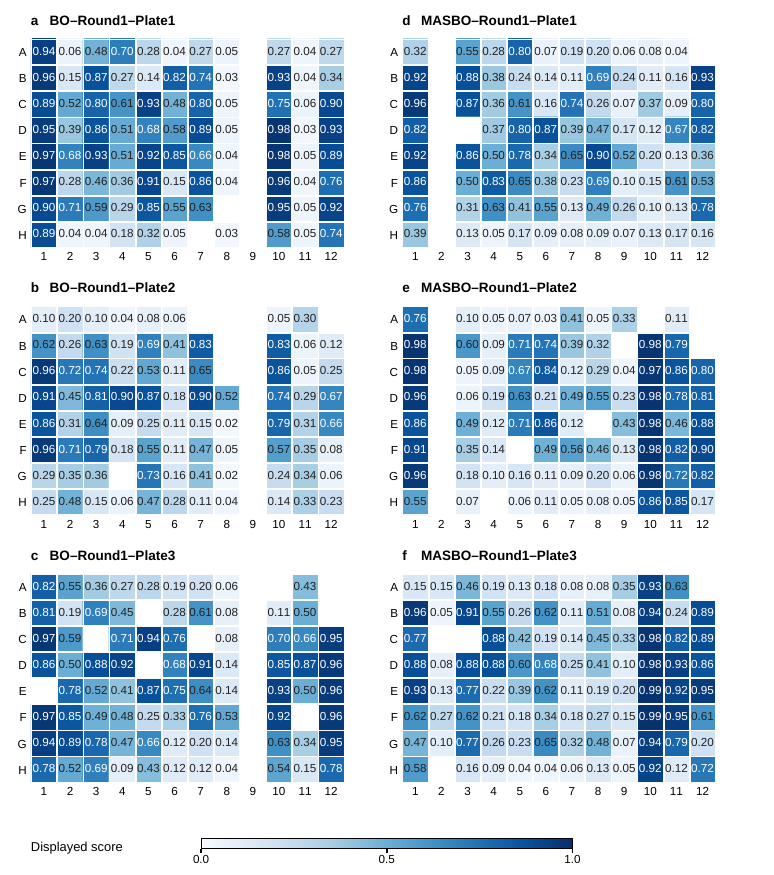}
\caption{\textbf{Plate-level phenotype scores in optimization round 1.} \textbf{a--c}, BO plates; \textbf{d--f}, MASBO plates. Numbers show original well scores on a shared 0--1 colour scale; white cells indicate masked or missing values. Condition-level failure-zero assignment is described in Methods.}
\label{fig:raw_heatmaps_r1}
\end{figure}

\begin{figure}[!htbp]
\centering
\includegraphics[width=\textwidth]{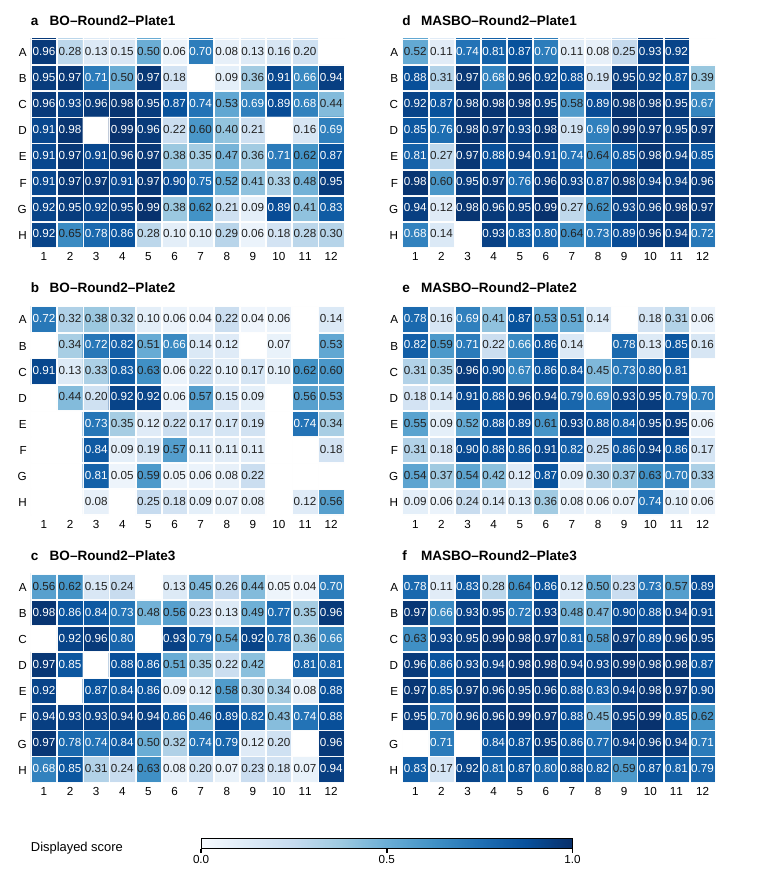}
\caption{\textbf{Plate-level phenotype scores in optimization round 2.} \textbf{a--c}, BO plates; \textbf{d--f}, MASBO plates. Numbers show original well scores on a shared 0--1 colour scale; white cells indicate masked or missing values.}
\label{fig:raw_heatmaps_r2}
\end{figure}

\begin{figure}[!htbp]
\centering
\includegraphics[width=\textwidth]{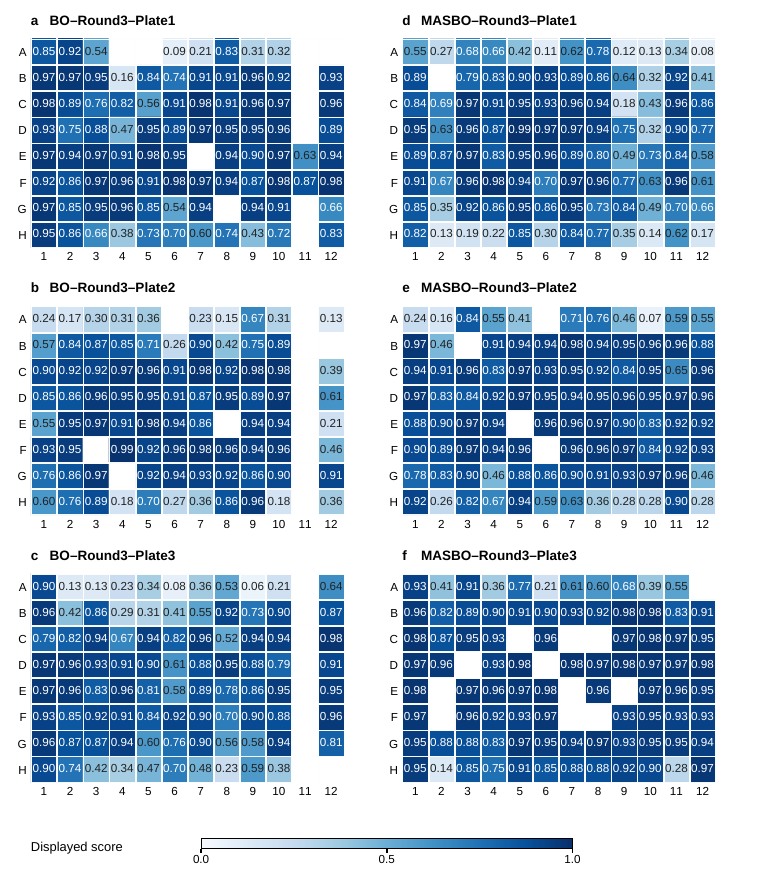}
\caption{\textbf{Plate-level phenotype scores in optimization round 3.} \textbf{a--c}, BO plates; \textbf{d--f}, MASBO plates. Numbers show original well scores on a shared 0--1 colour scale; white cells indicate masked or missing values. BO column 11 retains its original well scores here but is assigned zero as an experimental failure in condition-level analyses.}
\label{fig:raw_heatmaps_r3}
\end{figure}

Figs.~\ref{fig:raw_heatmaps_r1}--\ref{fig:raw_heatmaps_r3} show scores for all 18 plates on a common 0--1 colour scale. White cells denote masked or missing values, without imputation. BO round 1 W09, MASBO round 1 W02 and BO round 3 W11 are assigned zero as failures in the aggregate analyses. The heatmaps show the available scores from the original reports.

Complete condition-level data are provided in Source Data 2.
\input{tables/tab_opt_plate_layout}
\FloatBarrier

\FloatBarrier
\subsection{Factor space and sensitivity analyses}
\label{sec:app_c5}
\begin{figure}[htbp]
\centering
\includegraphics[width=\textwidth]{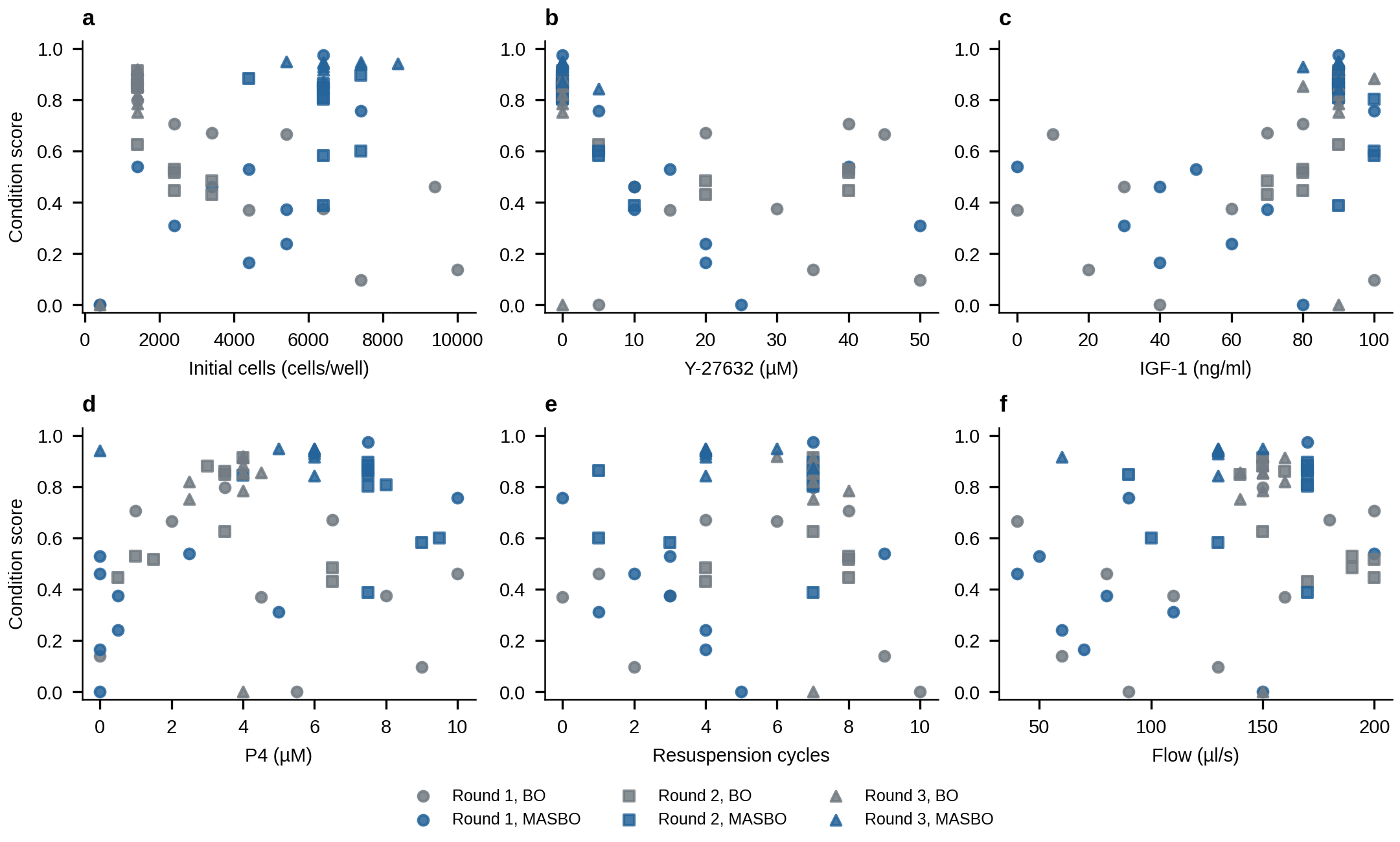}
\caption{\textbf{Observed culture conditions within the six-factor search space.} \textbf{a--f}, Scores for the same 60 conditions plotted against the six indicated factors (10 conditions per method and round). Scores are aggregated as in Fig.~\ref{fig:feedback}f. Grey and blue denote BO and MASBO; circles, squares and triangles denote rounds 1--3, respectively.}
\label{fig:ed7_factor_space}
\end{figure}

Fig.~\ref{fig:ed7_factor_space} shows factor settings and measured scores for all 60 conditions across the three rounds. These multivariable observations do not establish factor effects or causal interactions.

In round 3, averaging all three plates gave MASBO and BO means of 0.884 and 0.775, with medians of 0.913 and 0.848. Six of ten MASBO conditions and three of ten BO conditions reached 0.9. This analysis retained MASBO--Round3--Plate1 despite its substantial differences from MASBO--Round3--Plate2 and MASBO--Round3--Plate3. Excluding W11 from both methods gave nine-condition means of 0.924 and 0.849. Excluding BO W11 solely from the correlation calculation reduced the selected-pair correlation from 0.985 to 0.796 without changing the pair. The primary analysis retained failures as zero scores; candidate-wise three-plate medians and alternative-pair results are provided in the source data.

\FloatBarrier

\subsection{Retrospective optimization benchmarks}
\label{sec:app_c7}
\label{app:opt_experiment}
\subsubsection{Iris classification}
\label{sec:iris_retrospective}

We compared BO, LLAMBO and MASBO for hyperparameter optimization on the Iris dataset \cite{fisher1936iris}, with classification accuracy as the objective to be maximized. The evaluation covered random forest, decision tree, support vector machine, AdaBoost and a multilayer perceptron trained with stochastic gradient descent. Each classifier underwent 25 evaluations, comprising five initial configurations and 20 subsequent optimization steps. This offline benchmark was independent of the robotic retinal-organoid culture experiments.

Mean best accuracy is the average, across the five classifiers, of the highest accuracy attained within 25 evaluations:
\begin{equation}
\mathrm{MBP}_{m} = \frac{1}{5}\sum_{a=1}^{5}
\max_{1\leq t\leq25} y_{a,m,t}.
\label{eq:iris_mean_best_performance}
\end{equation}
Here, $y_{a,m,t}$ denotes the accuracy obtained by method $m$ on classifier $a$ at evaluation $t$.

\begin{center}
\begin{minipage}{\textwidth}
\captionsetup{type=table}
\begin{threeparttable}
\tablebodyfont
\centering
\caption{\textbf{Hyperparameter optimization on Iris.} Mean best accuracy averages the best accuracy attained for each of five classifiers. Bold indicates the best value for each metric; arrows indicate the preferred direction.}
\label{tab:iris_retrospective_selected}
\begin{tabular*}{\textwidth}{@{\extracolsep{\fill}}lcc@{}}
\toprule
Method & \shortstack{Mean best\\accuracy $\uparrow$} &
\shortstack{Final global\\regret $\downarrow$} \\
\midrule
BO     & 0.9050          & 0.0683 \\
LLAMBO & 0.9583          & 0.0150 \\
MASBO  & \textbf{0.9617} & \textbf{0.0000} \\
\bottomrule
\end{tabular*}
\end{threeparttable}
\end{minipage}
\end{center}

MASBO achieved a mean best accuracy of 0.9617, compared with 0.9050 for BO and 0.9583 for LLAMBO (Table~\ref{tab:iris_retrospective_selected}). The differences were 5.67 percentage points relative to BO and 0.34 percentage points relative to LLAMBO.

\subsubsection{iPSC--RPE differentiation}
\label{sec:ipsc_retrospective}

We also evaluated offline optimization using 143 experimental records of induced pluripotent stem cell-derived retinal pigment epithelium (iPSC--RPE) differentiation \cite{kanda2022robotic}. The search space comprised seven variables: preconditioning concentration (PC), preconditioning period (PP), detachment trypsin period (DP), detachment pipetting strength (DS) and detachment pipetting length (DL). The remaining variables were KnockOut Serum Replacement period (KP) and three-supplements period (3P). The objective was the pigmented-cell area fraction on day 34.

The offline evaluation framework supported dataset lookup and nearest-neighbour-based evaluation of proposed configurations. The benchmark compared BO, LLAMBO and MASBO; corresponding run configurations and identifiers are described in the source-data documentation.

Evaluation used normalized cumulative regret, area under the regret curve (AUC), sample mean regret and the proportion of high-performing samples. Lower values indicate better performance for the three regret metrics, whereas a higher proportion of high-performing samples is preferred.

\begin{center}
\begin{minipage}{\textwidth}
\captionsetup{type=table}
\begin{threeparttable}
\tablebodyfont
\centering
\caption{\textbf{Offline optimization of iPSC--RPE differentiation.} Four optimization metrics are compared across BO, LLAMBO and MASBO. Bold indicates the best value for each metric; arrows indicate the preferred direction.}
\label{tab:ipsc_retrospective_selected}
\begin{tabular}{@{}p{0.48\linewidth}ccc@{}}
\toprule
Metric & BO & LLAMBO & MASBO \\
\midrule
Normalized cumulative regret $\downarrow$
  & 8.7927 & 3.7850 & \textbf{3.7708} \\
Regret AUC $\downarrow$
  & 8.2240 & 3.5818 & \textbf{3.5724} \\
Sample mean regret $\downarrow$
  & 0.3983 & 0.1715 & \textbf{0.1708} \\
High-performance ratio $\uparrow$
  & 0.2150 & 0.2600 & \textbf{0.6100} \\
\bottomrule
\end{tabular}
\end{threeparttable}
\end{minipage}
\end{center}

MASBO and LLAMBO had similar values across the three regret metrics, with both below BO (Table~\ref{tab:ipsc_retrospective_selected}). The proportion of high-performing samples differed more markedly: 61.0\% for MASBO, 26.0\% for LLAMBO and 21.5\% for BO.

%% file: tables/tab_opt_round_summary.tex
\begin{table*}[htbp]
\centering
\scriptsize
\caption{Round-wise optimization outcomes. All statistics are calculated from the final score obtained using the highest-Pearson-r plate pair.}
\label{tab:optimization_round_summary}
\resizebox{\textwidth}{!}{%
\begin{tabular}{lllllllllll}
\toprule
Round & Algorithm & Selected pair & Mean & Median & \shortstack{Maximum\\score} & \shortstack{Top-3\\mean} & \shortstack{Fraction\\$\geq 0.8$} & \shortstack{Fraction\\$\geq 0.9$} & Failures & High-noise \\
\midrule
\multirow{2}{*}{Round 1} & BO & \shortstack[l]{BO--Round1--Plate1\\BO--Round1--Plate2} & 0.429 & 0.418 & 0.798 & 0.725 & 0.000 & 0.000 & 1 & 1 \\
 & MASBO & \shortstack[l]{MASBO--Round1--Plate2\\MASBO--Round1--Plate3} & 0.435 & 0.418 & 0.974 & 0.757 & 0.100 & 0.100 & 1 & 2 \\
\midrule
\multirow{2}{*}{Round 2} & BO & \shortstack[l]{BO--Round2--Plate1\\BO--Round2--Plate3} & 0.654 & 0.578 & 0.915 & 0.885 & 0.400 & 0.100 & 0 & 0 \\
 & MASBO & \shortstack[l]{MASBO--Round2--Plate1\\MASBO--Round2--Plate2} & 0.752 & 0.826 & 0.897 & 0.881 & 0.700 & 0.000 & 0 & 0 \\
\midrule
\multirow{2}{*}{Round 3} & BO & \shortstack[l]{BO--Round3--Plate2\\BO--Round3--Plate3} & 0.764 & 0.855 & 0.920 & 0.906 & 0.700 & 0.200 & 1 & -- \\
 & MASBO & \shortstack[l]{MASBO--Round3--Plate2\\MASBO--Round3--Plate3} & 0.923 & 0.941 & 0.949 & 0.949 & 1.000 & 0.800 & 0 & -- \\
\bottomrule
\end{tabular}%
}
\par\vspace{0.35em}\raggedright\footnotesize Failures are retained as score 0. High-noise is a descriptive QC label and is not an exclusion from the reported candidate denominator. Round-3 high-noise categories were not assigned. Plate identifiers specify the algorithm, optimization round and independently cultured batch plate; plate-level records are provided in Source Data 2.
\end{table*}

%% file: tables/tab_opt_plate_layout.tex
\begin{table*}[!htbp]
\centering
\small
\caption{Plate layout and score aggregation for organoid optimization. W02--W11 identify conditions, whereas a coordinate such as B2 identifies an individual well.}
\label{tab:opt_plate_layout}
\begin{tabularx}{\textwidth}{@{}lX@{}}
\toprule
Region or unit & Role in analysis \\
\midrule
Columns 1 and 12 & Control columns (W01/W12); excluded from candidate scores. \\
Rows A and H & Edge rows; excluded from candidate scores. \\
Rows B--G, columns 2--11 & Candidate region. Each column corresponds to one condition, with up to six technical-replicate wells per plate. \\
Blank or invalid measurements & Excluded from the within-plate mean; not converted to zero. \\
Plate1, Plate2 and Plate3 & Independent biological batches, each represented by one plate. Full identifiers specify algorithm, optimization round and plate (for example, BO--Round1--Plate1). \\
Final condition score & Average valid wells within each plate, then equally average the condition means from the highest-Pearson-correlation plate pair. \\
Recorded condition failures & BO round 1 W09, MASBO round 1 W02 and BO round 3 W11 retain score 0 at the condition level, rather than averaging any remaining well observations. Original well records are retained. \\
High-noise conditions & Retained in the analysis; the QC label does not exclude a condition. \\
\bottomrule
\end{tabularx}
\end{table*}

%% file: appendices/appendix_d_drug_repurposing.tex
\section{Drug repurposing and experimental evaluation}
\label{app:drug_repurposing}
\setcounter{figure}{0}
\setcounter{table}{0}

\FloatBarrier
\subsection{Graph-guided candidate nomination}
\label{sec:app_d1}
The drug-repurposing workflow first parses the disease, experimental model and study objective. Pathogenesis literature and supplied background support candidate genes or proteins, which are standardized against the knowledge graph. Only mapped targets continue. A relation-filter agent ranks mechanistic centrality and retains a target set for association retrieval or prediction.

The one-hop branch retrieves drugs directly linked to mapped targets, whereas the prediction branch proposes unobserved associations. Candidate origins are listed in Table~\ref{tab:drug_evidence_portfolio}.

In the prediction branch, three saved graph neural-network model instances score drug--protein associations for each retained target. Each model contributes its highest-ranked predictions; only drugs recovered by at least two models are eligible. Associations already present as known drug--protein edges are removed, including global removal across all mapped disease targets, and duplicate drug--target pairs are collapsed. In the frozen configuration, prediction is capped at 50 associations after consensus and known-edge filtering. Scores are min--max normalized within each target before aggregation by drug to reduce target-specific scale effects. For drug $d$, the structural score is
\[
H_d=0.6\,\overline{s}^{\,\mathrm{norm}}_d
+0.3\,s^{\mathrm{norm,max}}_d
+0.1\,c_d,
\]
where $c_d$ is the fraction of filtered targets linked to $d$. Candidates are first ordered by $H_d$ and coverage. An LLM graph-ranking stage then evaluates target relevance and biological plausibility while retaining target identifiers, paths and structural scores.

When graph candidates are available, they become traceable hypotheses and the generic free-form generation agent is bypassed. Each hypothesis then undergoes four recurrent operations. First, the system critiques mechanistic weaknesses and task mismatch. Second, it retrieves supporting, contradictory or mixed evidence and records references and gaps. Third, evidence-aware evolution preserves parent--child lineage. Finally, weighted ranking combines novelty, plausibility, testability and alignment with the objective \cite{gottweis2026coscientist,internagent2025}. Iteration records retain drugs, target paths, critiques, evidence, queries, gaps, evolved text and scores.

\FloatBarrier
\subsection{Candidate evidence and evidence-chain assessment}
\label{sec:app_d2}
\label{sec:app_d3}
\label{app:drug_evidence_chains}

Evidence was reviewed across rounds within one parent-linked lineage per candidate, without pooling sources from different runs or branches. Sources were selected for drug, mechanism and model relevance rather than restricted to the final round. Table~\ref{tab:drug_evidence_portfolio} separates archived system hypotheses from retrieved findings, including drug-specific evidence and findings that constrain mechanistic interpretation.

Candidates were classified as \emph{prior-supported} when the tested relationship was substantially established, or \emph{exploratory} when it required non-obvious reasoning across sources. These categories describe the literature evidence supporting the drug candidates.

\input{tables/tab_drug_evidence_portfolio}

Table~\ref{tab:drug_preserved_sources} lists 21 sources selected across the six candidate lineages, with retrieval rounds, study models, findings and relevance.
\FloatBarrier
\input{tables/tab_drug_preserved_sources}

\FloatBarrier
\subsection{Drug stocks and vehicles}
\label{sec:app_d4}

\noindent\begin{minipage}{\textwidth}
\centering
\small
\captionof{table}{\textbf{Drug stock formulations and calculated final DMSO fractions.} Stocks were serially diluted in culture medium without adding further solvent. The last three columns give stock-derived DMSO percentages at each final drug concentration. Values were calculated as $100\times C_{\mathrm{final}}/C_{\mathrm{stock}}$, with concentrations expressed in the same units. Water-based stocks contributed no DMSO. DMSO fractions were not matched across drugs or doses, and matched vehicle controls were not included. MK-801 served as the positive control.}
\label{tab:drug_stock_vehicles}
\begin{tabularx}{\textwidth}{@{}Xrlrrr@{}}
\toprule
 & & & \multicolumn{3}{c}{DMSO (\%, v/v) at drug concentration} \\
Compound & Stock (mM) & Solvent & 1~$\mu$M & 10~$\mu$M & 100~$\mu$M \\
\midrule
Ifenprodil & 20 & DMSO & 0.005 & 0.05 & 0.5 \\
Milnacipran HCl & 20 & Sterile water & 0 & 0 & 0 \\
Stiripentol & 10 & DMSO & 0.01 & 0.1 & 1 \\
Phenytoin & 50 & DMSO & 0.002 & 0.02 & 0.2 \\
Memantine HCl & 50 & DMSO & 0.002 & 0.02 & 0.2 \\
Agmatine sulfate & 100 & Sterile water & 0 & 0 & 0 \\
MK-801 & 50 & DMSO & 0.002 & 0.02 & 0.2 \\
\bottomrule
\end{tabularx}
\end{minipage}
\par

\FloatBarrier
\subsection{Preliminary screening and follow-up}
\label{sec:app_d5}
\begin{figure}[htbp]
\centering
\makebox[\textwidth][c]{\includegraphics[width=158mm]{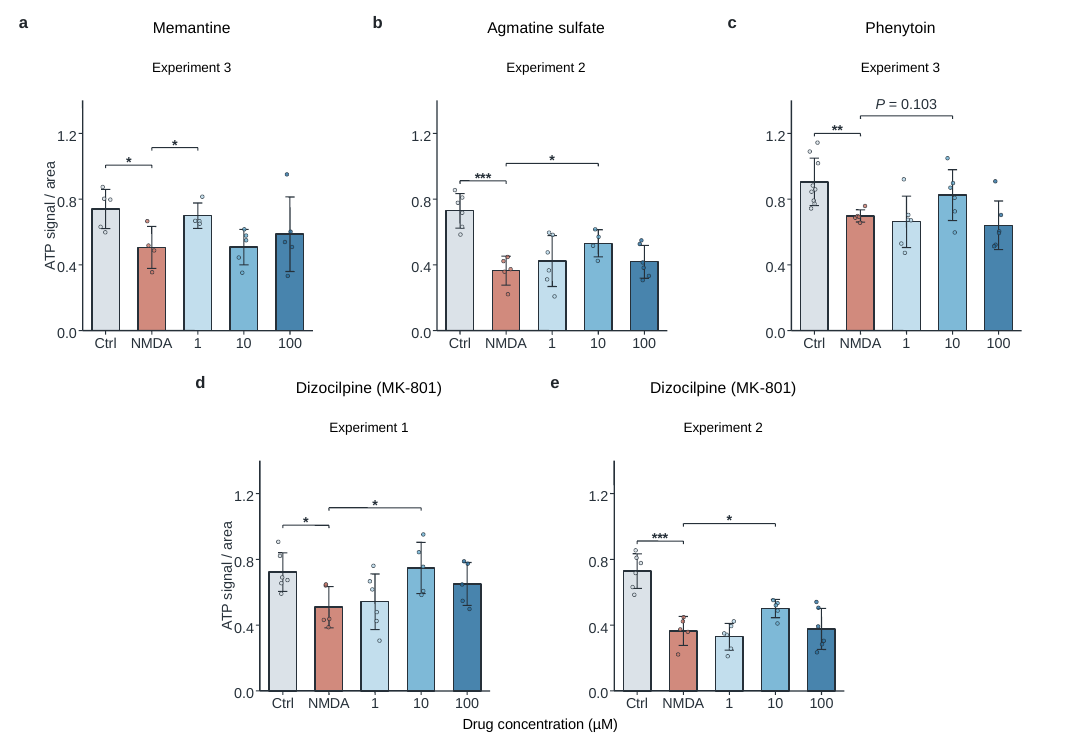}}
\caption{\textbf{Additional drug experiments during NMDA exposure.} \textbf{a--c}, Memantine experiment 3, agmatine experiment 2 and phenytoin experiment 3; \textbf{d,e}, MK-801 experiments 1 and 2. Numbers identify datasets. Panels \textbf{b,e} share Ctrl and NMDA controls; panel \textbf{d} shares its control groups with Mem, Agm and Phe in Fig.~\ref{fig:drug}e. Dots denote organoids; bars show means with SD error bars for area-normalized ATP signals. Group sizes are listed in Table~\ref{tab:drug_response_summary}. Ctrl, normal medium; NMDA, NMDA alone; drug groups received simultaneous NMDA exposure. Bracketed comparisons use two-sided Welch t-tests with unadjusted P values: $^{*}P<0.05$, $^{**}P<0.01$, $^{***}P<0.001$. Other experiments appear in Fig.~\ref{fig:drug}d,e.}
\label{fig:drug_independent_experiments}
\end{figure}

The 15 compound-specific datasets comprise six preliminary screens (Fig.~\ref{fig:drug}d), four follow-up experiments (Fig.~\ref{fig:drug}e) and five additional experiments (Fig.~\ref{fig:drug_independent_experiments}). Dataset identifiers do not denote independent culture batches. Table~\ref{tab:drug_response_summary} summarizes individual experiments and sample sizes for all six candidates and MK-801.

\FloatBarrier
\clearpage
\input{tables/tab_drug_response_summary}

%% file: tables/tab_drug_evidence_portfolio.tex
\begingroup
\footnotesize
\setlength{\tabcolsep}{3pt}
\renewcommand{\arraystretch}{1.12}
\begin{longtable}{@{}>{\raggedright\arraybackslash}p{0.16\textwidth}>{\raggedright\arraybackslash}p{0.37\textwidth}>{\raggedright\arraybackslash}p{\dimexpr0.47\textwidth-4\tabcolsep\relax}@{}}
\caption{\textbf{System-generated hypotheses and retrieved evidence for six BioMARS candidates.} Hypotheses summarize archived outputs from one selected parent-linked lineage per candidate. Evidence is selected across all rounds of that lineage, prioritizing drug-specific findings, relevant mechanisms and retinal models while retaining findings that constrain interpretation. Proposed mechanisms are distinguished from literature findings.}\label{tab:drug_evidence_portfolio}\\
\toprule
Candidate/\newline origin & System-generated hypothesis & Retrieved evidence from the same lineage \\
\midrule
\endfirsthead
\multicolumn{3}{l}{Table~\thetable\ (continued)}\\
\toprule
Candidate/\newline origin & System-generated hypothesis & Retrieved evidence from the same lineage \\
\midrule
\endhead
\midrule
\multicolumn{3}{r}{\textit{Continued on the next page}}\\
\endfoot
\bottomrule
\endlastfoot
Ifenprodil\newline One-hop link & Inhibition of GluN2B-containing NMDARs would reduce intracellular calcium overload and improve viability during NMDA exposure in retinal organoids. & Receptor pharmacology and developing rat cortical neurons support GluN2B-dependent antagonism and subtype-dependent ifenprodil sensitivity (rounds 2--5) \cite{williams2001ifenprodil,kew1998evidence9482779}. These findings ground the receptor component of the hypothesis; organoid protection is an extrapolation. \\
\addlinespace[4pt]
Milnacipran HCl\newline One-hop link & Non-competitive NMDAR antagonism would reduce GRIN1-associated calcium influx and subsequent inflammation. The system proposed IL-6 and TNF-$\alpha$ measurements to test this response. & Retrieved literature reports non-competitive NMDAR antagonism by milnacipran itself \cite{shuto2001evidence11325012} and NMDA-channel blockade and neuroprotection by conformationally restricted analogues (rounds 1--6) \cite{shuto1998evidence9719604,noguchi1999milnacipran}. These findings support an NMDAR-related pharmacological rationale, but do not establish the proposed anti-inflammatory mechanism in organoids. \\
\addlinespace[4pt]
Agmatine sulfate\newline One-hop link\newline GRIN1 & Competitive antagonism at GRIN2B and activation of Nrf2 were proposed to reduce calcium influx and oxidative stress during NMDA exposure. & Polyamine-site binding and spinal NMDA-evoked calcium transients support receptor modulation (rounds 1--5) \cite{gibson2002agmatine,xie2025evidence39969272}. Rotenone-lesioned rats provide Nrf2-related evidence (rounds 1--5) \cite{azar2022agmatinenrf2}; a retinal-neuroinflammation study reports protection in mice (round 5) \cite{monu2025agmatineretina}. Polyamine-site evidence does not establish competitive GRIN2B antagonism. \\
\addlinespace[4pt]
Stiripentol\newline Predicted link & GRIN2A/2B modulation was refined into a hypothesis of enhanced GRIN2B-mediated signalling, improved mitochondrial dynamics and reduced oxidative stress in retinal organoids. & A drug-specific study reports neuronal protection and reduced sodium/calcium entry, but no detectable binding affinity for the tested glutamate receptors (round 4) \cite{verleye2016stiripentol}. Retinal-organoid evidence links mitochondrial oxidative stress to RGC death (round 5) \cite{hirata2025evidence40651649}. Neither establishes the proposed enhancement of GRIN2B signalling. \\
\addlinespace[4pt]
Phenytoin\newline Predicted link & GRIN2B/2D modulation would reduce reactive oxygen species and improve mitochondrial respiration during NMDA exposure. Proposed readouts were oxygen consumption and malondialdehyde levels. & Phenytoin produced modest electrocorticographic correction in a GRIN2D-variant mouse model (rounds 1--5) \cite{yam2026evidence40277233}. A neuronal-injury study links GluN2B signalling to mitophagy with THSG, a different compound (round 2) \cite{liu2024thsg}. This supports a pathway rationale, not direct receptor engagement by phenytoin. \\
\addlinespace[4pt]
Memantine\newline One-hop link\newline GRIN1 & Selective inhibition of extrasynaptic NMDARs and GRIN1-dependent desensitization were proposed to reduce calcium overload and NMDA-induced organoid injury. & Retrieved studies report protection against NMDA-induced retinal degeneration (round 1) \cite{kobayashi2017evidence28253402}, calcium-dependent NMDAR desensitization (rounds 1--4) \cite{glasgow2017evidence28877967}, and autophagy in ARPE-19 cells (rounds 3--4) \cite{lee2023evidence37641807}. Organoid-specific receptor selectivity remains part of the hypothesis. \\
\end{longtable}
\noindent Selected lineages cover rounds 1--5 for ifenprodil, agmatine, stiripentol and phenytoin; rounds 1--6 for milnacipran; and rounds 1--4 for memantine. Origins denote existing graph edges or predicted associations, separately from mechanistic validation. Study-level findings are listed in Table~\ref{tab:drug_preserved_sources}.
\endgroup

%% file: tables/tab_drug_preserved_sources.tex
\begingroup
\footnotesize
\setlength{\tabcolsep}{3pt}
\renewcommand{\arraystretch}{1.12}
\begin{longtable}{@{}>{\raggedright\arraybackslash}p{0.16\textwidth}>{\raggedright\arraybackslash}p{0.10\textwidth}>{\raggedright\arraybackslash}p{0.36\textwidth}>{\raggedright\arraybackslash}p{\dimexpr0.38\textwidth-6\tabcolsep\relax}@{}}
\caption{\textbf{Literature evidence retrieved across candidate-refinement rounds.} Twenty-one sources are grouped by candidate and drawn from the same lineages as Table~\ref{tab:drug_evidence_portfolio}. D identifies a source within a candidate; R identifies refinement rounds containing that source. Study findings are reported separately from model inferences.}\label{tab:drug_preserved_sources}\label{tab:mem_preserved_sources}\\
\toprule
Candidate/\newline source / rounds & Ref. & Study model and principal finding & Relevance to this study \\
\midrule
\endfirsthead
\multicolumn{4}{l}{Table~\thetable\ (continued)}\\
\toprule
Candidate/\newline source / rounds & Ref. & Study model and principal finding & Relevance to this study \\
\midrule
\endhead
\midrule
\multicolumn{4}{r}{\textit{Continued on the next page}}\\
\endfoot
\bottomrule
\endlastfoot
Ifenprodil\newline D1\newline R2--5 & \cite{williams2001ifenprodil} & Receptor pharmacology: selective antagonism of NR2B/GluN2B-containing NMDARs. & Direct pharmacological support; retinal organoids were not examined. \\
\addlinespace[3pt]
Ifenprodil\newline D2\newline R2--5 & \cite{kew1998evidence9482779} & Developing rat cortical neurons: NMDAR subtypes differ in ifenprodil sensitivity. & Supports receptor-subtype dependence; extrapolation from developing cortical neurons is required. \\
\addlinespace[3pt]
Ifenprodil\newline D3\newline R2--5 & \cite{kong2024evidence38685368} & A mouse PTSD model: Cheonwangbosimdan reduced PTSD-like behaviours through GluN2B-containing NMDAR antagonism. & Supports the target mechanism using a different compound and disease model. \\
\addlinespace[3pt]
Milnacipran HCl\newline D1\newline R1--6 & \cite{shuto1998evidence9719604} & Conformationally restricted milnacipran analogues show NMDA-channel-blocking activity. & Analogue-based support, rather than direct evidence for the parent drug. \\
\addlinespace[3pt]
Milnacipran HCl\newline D2\newline R1--6 & \cite{shuto2001evidence11325012} & Milnacipran itself has non-competitive NMDAR-antagonist activity; its conformationally restricted analogue PPDC is approximately 30-fold more potent. & Supports NMDAR pharmacology of the parent drug and analogue development; no retinal model was tested. \\
\addlinespace[3pt]
Milnacipran HCl\newline D3\newline R1--6 & \cite{noguchi1999milnacipran} & Conformationally restricted analogues block open NMDA channels and protect against NMDA-induced neurotoxicity. & Analogue potency and dose cannot be transferred directly to milnacipran. \\
\addlinespace[3pt]
Agmatine sulfate\newline D1\newline R1--5 & \cite{xie2025evidence39969272} & Mouse spinal dorsal horn: agmatine inhibition of NMDA-evoked calcium transients required an intact NMDAR--PSD95--nNOS pathway and persisted after GluN2B knockdown. & Direct drug-level support in non-retinal tissue; does not establish GluN2B dependence. \\
\addlinespace[3pt]
Agmatine sulfate\newline D2\newline R1--5 & \cite{gibson2002agmatine} & Radioligand binding supports agmatine activity at the NMDAR polyamine site. & Supports polyamine-site pharmacology, not competitive GRIN2B antagonism. \\
\addlinespace[3pt]
Agmatine sulfate\newline D3\newline R3,5 & \cite{kotagale2019evidence31063707} & Review of neuroprotective effects of agmatine across experimental settings. & Broad background support rather than a primary test in the target model. \\
\addlinespace[3pt]
Agmatine sulfate\newline D4\newline R1--5 & \cite{azar2022agmatinenrf2} & Rotenone-lesioned rats: agmatine reduced NMDAR expression and oxidative/inflammatory markers, with increased Nrf2-associated antioxidant responses. & Drug-specific support for the Nrf2 component in non-retinal tissue; not evidence of competitive GRIN2B antagonism. \\
\addlinespace[3pt]
Agmatine sulfate\newline D5\newline R5 & \cite{monu2025agmatineretina} & Mouse retinal neuroinflammation: agmatine reduced inflammation and protected RGC function. Protection against oxidative stress was also observed in 661W cells. & Drug-specific retinal evidence from an inflammatory injury model, distinct from NMDA-challenged organoids. \\
\addlinespace[3pt]
Stiripentol\newline D4\newline R4 & \cite{verleye2016stiripentol} & Rat cortical neuronal--astroglial cultures: stiripentol protected against oxygen--glucose deprivation or glutamate injury and reduced veratridine-evoked sodium and NMDA-evoked calcium entry. & Direct drug evidence. Binding assays found no detectable affinity for the tested glutamate receptors (IC$_{50}>100~\mu$M); the study does not establish enhanced GRIN2B signalling. \\
\addlinespace[3pt]
Stiripentol\newline D1\newline R5 & \cite{zhang2026evidence41811034} & NFI23, a GluN2B-NMDAR antagonist, was investigated for ischaemic stroke. & Supports target feasibility; the tested compound was NFI23, not stiripentol. \\
\addlinespace[3pt]
Stiripentol\newline D2\newline R5 & \cite{hirata2025evidence40651649} & Human iPSC-derived retinal organoids: hypoxia links mitochondrial oxidative stress and calpain activation to RGC death. & Relevant retinal-organoid pathway evidence; no stiripentol treatment was studied. \\
\addlinespace[3pt]
Stiripentol\newline D3\newline R5 & \cite{haque2021evidence33285471} & Primary hippocampal neurons: stigmasterol modulates mitophagy during hypoxia/reoxygenation injury. & Supports a downstream mechanism with a different drug and model. \\
\addlinespace[3pt]
Phenytoin\newline D1\newline R1--5 & \cite{yam2026evidence40277233} & A GRIN2D pathogenic-variant mouse model: phenytoin produced modest electrocorticographic correction. & Direct drug evidence, but neither the model nor endpoint represents retinal excitotoxic injury. \\
\addlinespace[3pt]
Phenytoin\newline D3\newline R2 & \cite{liu2024thsg} & Rat cerebral ischaemia/reperfusion and neuronal oxygen--glucose deprivation/reoxygenation: THSG promoted GluN2B--CaMKII--ERK1/2-associated PINK1--PARK2 mitophagy. GluN2B inhibition attenuated protection. & Supports a receptor--mitochondrial pathway in neuronal injury using THSG, not phenytoin; it does not demonstrate phenytoin--GRIN binding. \\
\addlinespace[3pt]
Phenytoin\newline D2\newline R5 & \cite{iwata2022evidence35435002} & D-alanine in kidney injury: links among NMDAR signalling, reactive oxygen species and mitochondrial potential. & Pathway-level support using D-alanine and a non-retinal model. \\
\addlinespace[3pt]
Memantine\newline D1\newline R1 & \cite{kobayashi2017evidence28253402} & NMDA-induced retinal injury: memantine reduced retinal degeneration. & Direct retinal protection evidence; organoid-specific efficacy and mechanism require separate testing. \\
\addlinespace[3pt]
Memantine\newline D2\newline R1--4 & \cite{glasgow2017evidence28877967} & Receptor experiments support calcium-dependent modulation of NMDAR desensitization by memantine. & Does not establish GRIN1-specific or extrasynaptic-selective protection in retinal organoids. \\
\addlinespace[3pt]
Memantine\newline D3\newline R3--4 & \cite{lee2023evidence37641807} & ARPE-19 cells: memantine and ifenprodil promote autophagy-dependent A2E removal. & Retinal-cell support with a model and endpoint distinct from NMDA-induced organoid injury. \\
\end{longtable}
\endgroup

%% file: tables/tab_drug_response_summary.tex
\begingroup
\footnotesize
\setlength{\tabcolsep}{4pt}
\begin{longtable}{llrrrrr}
\caption{\textbf{ATP viability measurements across the reported drug experiments.} Means and SDs summarize ATP luminescence divided by projected organoid area; $n$ denotes the number of organoids assayed. Drug-group changes are calculated relative to the displayed NMDA-group mean, without pooling datasets. Ifenprodil, stiripentol and milnacipran hydrochloride each have one preliminary screen. The three experiments per selected candidate include one preliminary screen and two subsequent experiments. Mem 2, Agm 3 and Phe 2 are preliminary-screen datasets; numbers do not indicate chronological order. Drug-treated and control groups in Agm 3, and drug-treated and NMDA reference groups in Phe 2, came from different batches; these comparisons are descriptive only. Ctrl denotes normal medium. MK-801 is a separate positive control, not one of the six nominated candidates. This table summarizes descriptive statistics; tests are specified in Methods and the corresponding figure legends.}\label{tab:drug_response_summary}\\
\toprule
Compound/experiment & Group & $n$ & Mean & SD & Change (\%) & Fig. 4 panel \\
\midrule
\endfirsthead
\multicolumn{7}{l}{\textit{Continued from the previous page}}\\
\toprule
Compound/experiment & Group & $n$ & Mean & SD & Change (\%) & Fig. 4 panel \\
\midrule
\endhead
\midrule
\multicolumn{7}{r}{\textit{Continued on the next page}}\\
\endfoot
\bottomrule
\endlastfoot
Ifenprodil 1 & Ctrl & 6 & 0.455 & 0.109 & -- & d \\
 & NMDA & 5 & 0.363 & 0.083 & -- &  \\
 & 1 $\mu$M & 8 & 0.297 & 0.107 & -18.1 &  \\
 & 10 $\mu$M & 8 & 0.318 & 0.104 & -12.3 &  \\
 & 100 $\mu$M & 8 & 0.297 & 0.101 & -18.2 &  \\
\addlinespace[2pt]
Stiripentol 1 & Ctrl & 6 & 0.838 & 0.130 & -- & d \\
 & NMDA & 7 & 0.552 & 0.120 & -- &  \\
 & 1 $\mu$M & 7 & 0.477 & 0.136 & -13.6 &  \\
 & 10 $\mu$M & 8 & 0.571 & 0.142 & +3.4 &  \\
 & 100 $\mu$M & 8 & 0.522 & 0.129 & -5.4 &  \\
\addlinespace[2pt]
Milnacipran HCl 1 & Ctrl & 6 & 0.838 & 0.130 & -- & d \\
 & NMDA & 7 & 0.552 & 0.120 & -- &  \\
 & 1 $\mu$M & 8 & 0.508 & 0.149 & -8.1 &  \\
 & 10 $\mu$M & 7 & 0.460 & 0.099 & -16.8 &  \\
 & 100 $\mu$M & 8 & 0.540 & 0.121 & -2.3 &  \\
\addlinespace[2pt]
MK-801 1 & Ctrl & 6 & 0.723 & 0.117 & -- &  \\
 & NMDA & 5 & 0.509 & 0.125 & -- &  \\
 & 1 $\mu$M & 6 & 0.542 & 0.169 & +6.5 &  \\
 & 10 $\mu$M & 5 & 0.748 & 0.156 & +46.9 &  \\
 & 100 $\mu$M & 5 & 0.651 & 0.130 & +27.8 &  \\
\addlinespace[2pt]
MK-801 2 & Ctrl & 6 & 0.729 & 0.105 & -- &  \\
 & NMDA & 5 & 0.365 & 0.088 & -- &  \\
 & 1 $\mu$M & 6 & 0.329 & 0.081 & -9.8 &  \\
 & 10 $\mu$M & 5 & 0.501 & 0.056 & +37.2 &  \\
 & 100 $\mu$M & 6 & 0.377 & 0.125 & +3.3 &  \\
\addlinespace[2pt]
MK-801 3 & Ctrl & 5 & 0.779 & 0.108 & -- & e \\
 & NMDA & 6 & 0.615 & 0.126 & -- &  \\
 & 1 $\mu$M & 6 & 0.764 & 0.158 & +24.3 &  \\
 & 10 $\mu$M & 6 & 0.842 & 0.118 & +37.0 &  \\
 & 100 $\mu$M & 5 & 0.782 & 0.197 & +27.2 &  \\
\addlinespace[2pt]
Memantine 1 & Ctrl & 6 & 0.723 & 0.117 & -- & e \\
 & NMDA & 5 & 0.509 & 0.125 & -- &  \\
 & 1 $\mu$M & 6 & 0.702 & 0.119 & +38.0 &  \\
 & 10 $\mu$M & 6 & 0.615 & 0.170 & +20.9 &  \\
 & 100 $\mu$M & 6 & 0.630 & 0.146 & +23.8 &  \\
\addlinespace[2pt]
Memantine 2 & Ctrl & 6 & 0.931 & 0.173 & -- & d \\
 & NMDA & 7 & 0.577 & 0.124 & -- &  \\
 & 1 $\mu$M & 8 & 0.748 & 0.160 & +29.6 &  \\
 & 10 $\mu$M & 8 & 0.618 & 0.184 & +7.1 &  \\
 & 100 $\mu$M & 8 & 0.683 & 0.183 & +18.3 &  \\
\addlinespace[2pt]
Memantine 3 & Ctrl & 5 & 0.740 & 0.119 & -- &  \\
 & NMDA & 4 & 0.506 & 0.127 & -- &  \\
 & 1 $\mu$M & 4 & 0.699 & 0.077 & +38.1 &  \\
 & 10 $\mu$M & 5 & 0.508 & 0.108 & +0.3 &  \\
 & 100 $\mu$M & 5 & 0.586 & 0.226 & +15.8 &  \\
\addlinespace[2pt]
Agmatine 1 & Ctrl & 6 & 0.723 & 0.117 & -- & e \\
 & NMDA & 5 & 0.509 & 0.125 & -- &  \\
 & 1 $\mu$M & 6 & 0.526 & 0.133 & +3.3 &  \\
 & 10 $\mu$M & 6 & 0.694 & 0.129 & +36.4 &  \\
 & 100 $\mu$M & 6 & 0.593 & 0.149 & +16.5 &  \\
\addlinespace[2pt]
Agmatine 2 & Ctrl & 6 & 0.729 & 0.105 & -- &  \\
 & NMDA & 5 & 0.365 & 0.088 & -- &  \\
 & 1 $\mu$M & 6 & 0.423 & 0.155 & +16.0 &  \\
 & 10 $\mu$M & 4 & 0.531 & 0.082 & +45.6 &  \\
 & 100 $\mu$M & 6 & 0.419 & 0.100 & +14.7 &  \\
\addlinespace[2pt]
Agmatine 3 & Ctrl & 9 & 0.904 & 0.144 & -- & d \\
 & NMDA & 5 & 0.698 & 0.037 & -- &  \\
 & 1 $\mu$M & 6 & 0.693 & 0.094 & -0.7 &  \\
 & 10 $\mu$M & 6 & 0.795 & 0.083 & +14.0 &  \\
 & 100 $\mu$M & 6 & 0.597 & 0.167 & -14.4 &  \\
\addlinespace[2pt]
Phenytoin 1 & Ctrl & 6 & 0.723 & 0.117 & -- & e \\
 & NMDA & 5 & 0.509 & 0.125 & -- &  \\
 & 1 $\mu$M & 6 & 0.527 & 0.106 & +3.6 &  \\
 & 10 $\mu$M & 5 & 0.647 & 0.086 & +27.2 &  \\
 & 100 $\mu$M & 6 & 0.564 & 0.109 & +10.8 &  \\
\addlinespace[2pt]
Phenytoin 2 & Ctrl & 6 & 0.729 & 0.105 & -- & d \\
 & NMDA & 5 & 0.365 & 0.088 & -- &  \\
 & 1 $\mu$M & 5 & 0.556 & 0.128 & +52.5 &  \\
 & 10 $\mu$M & 6 & 0.619 & 0.211 & +69.5 &  \\
 & 100 $\mu$M & 6 & 0.531 & 0.127 & +45.6 &  \\
\addlinespace[2pt]
Phenytoin 3 & Ctrl & 9 & 0.904 & 0.144 & -- &  \\
 & NMDA & 5 & 0.698 & 0.037 & -- &  \\
 & 1 $\mu$M & 6 & 0.662 & 0.156 & -5.2 &  \\
 & 10 $\mu$M & 6 & 0.824 & 0.154 & +18.1 &  \\
 & 100 $\mu$M & 6 & 0.641 & 0.148 & -8.2 &  \\
\addlinespace[2pt]
\end{longtable}
\endgroup

%% file: main.bbl
\begin{thebibliography}{75}
\expandafter\ifx\csname url\endcsname\relax
  \def\url#1{\burl{#1}}\fi
\expandafter\ifx\csname urlprefix\endcsname\relax\def\urlprefix{URL }\fi
\providecommand{\bibinfo}[2]{#2}
\providecommand{\eprint}[2][]{\url{#2}}
\providecommand{\doi}[1]{\url{https://doi.org/#1}}
\bibcommenthead

\bibitem{dettinger2022open}
\bibinfo{author}{Dettinger, P.~ \emph{et~al.}}
\newblock \bibinfo{title}{Open-source personal pipetting robots with live-cell incubation and microscopy compatibility}.
\newblock \emph{\bibinfo{journal}{Nature Communications}} \textbf{\bibinfo{volume}{13}}, \bibinfo{pages}{2999} (\bibinfo{year}{2022}).

\bibitem{kanda2022robotic}
\bibinfo{author}{Kanda, G.~N.~ \emph{et~al.}}
\newblock \bibinfo{title}{Robotic search for optimal cell culture in regenerative medicine}.
\newblock \emph{\bibinfo{journal}{eLife}} \textbf{\bibinfo{volume}{11}}, \bibinfo{pages}{e77007} (\bibinfo{year}{2022}).

\bibitem{tristan2021robotic}
\bibinfo{author}{Tristan, C.~A.~ \emph{et~al.}}
\newblock \bibinfo{title}{Robotic high-throughput biomanufacturing and functional differentiation of human pluripotent stem cells}.
\newblock \emph{\bibinfo{journal}{Stem Cell Reports}} \textbf{\bibinfo{volume}{16}}, \bibinfo{pages}{3076--3092} (\bibinfo{year}{2021}).

\bibitem{lan2025autobio}
\bibinfo{author}{Lan, Z.~\emph{et~al.}}
\newblock \bibinfo{title}{AutoBio: A simulation and benchmark for robotic automation in digital biology laboratory}.
\newblock Preprint at \url{https://arxiv.org/abs/2505.14030} (\bibinfo{year}{2025}).

\bibitem{hamm2024modular}
\bibinfo{author}{Hamm, J.~ \emph{et~al.}}
\newblock \bibinfo{title}{A Modular Robotic Platform for Biological Research: Cell Culture Automation and Remote Experimentation}.
\newblock \emph{Advanced Intelligent Systems} \textbf{6}, 2300566 (2024).

\bibitem{koniger2024rebia}
\bibinfo{author}{K{\"o}niger, L.~ \emph{et~al.}}
\newblock \bibinfo{title}{ReBiA---Robotic Enabled Biological Automation: 3D Epithelial Tissue Production}.
\newblock \emph{Advanced Science} \textbf{11}, 2406608 (2024).

\bibitem{king2004robotscientist}
\bibinfo{author}{King, R.~D.~ \emph{et~al.}}
\newblock \bibinfo{title}{Functional genomic hypothesis generation and experimentation by a robot scientist}.
\newblock \emph{\bibinfo{journal}{Nature}} \textbf{\bibinfo{volume}{427}}, \bibinfo{pages}{247--252} (\bibinfo{year}{2004}).

\bibitem{burger2020mobile}
\bibinfo{author}{Burger, B.~ \emph{et~al.}}
\newblock \bibinfo{title}{A mobile robotic chemist}.
\newblock \emph{\bibinfo{journal}{Nature}} \textbf{\bibinfo{volume}{583}}, \bibinfo{pages}{237--241} (\bibinfo{year}{2020}).

\bibitem{macleod2020selfdriving}
\bibinfo{author}{MacLeod, B.~P.~ \emph{et~al.}}
\newblock \bibinfo{title}{Self-driving laboratory for accelerated discovery of thin-film materials}.
\newblock \emph{\bibinfo{journal}{Science Advances}} \textbf{\bibinfo{volume}{6}}, \bibinfo{pages}{eaaz8867} (\bibinfo{year}{2020}).

\bibitem{szymanski2023alab}
\bibinfo{author}{Szymanski, N.~J.~ \emph{et~al.}}
\newblock \bibinfo{title}{An autonomous laboratory for the accelerated synthesis of inorganic materials}.
\newblock \emph{\bibinfo{journal}{Nature}} \textbf{\bibinfo{volume}{624}}, \bibinfo{pages}{86--91} (\bibinfo{year}{2023}).

\bibitem{odonoghue2023bioplanner}
\bibinfo{author}{O'Donoghue, O.~ \emph{et~al.}}
\newblock \bibinfo{title}{BioPlanner: Automatic Evaluation of LLMs on Protocol Planning in Biology}.
\newblock \emph{Proceedings of the 2023 Conference on Empirical Methods in Natural Language Processing}, 2676--2694 (2023).

\bibitem{qu2025crisprgpt}
\bibinfo{author}{Qu, Y.~ \emph{et~al.}}
\newblock \bibinfo{title}{CRISPR-GPT for agentic automation of gene-editing experiments}.
\newblock \emph{Nature Biomedical Engineering} \textbf{10}, 245--258 (2026).

\bibitem{boiko2023autonomous}
\bibinfo{author}{Boiko, D.~A.~, MacKnight, R.~, Kline, B.~ \& Gomes, G.~}
\newblock \bibinfo{title}{Autonomous chemical research with large language models}.
\newblock \emph{\bibinfo{journal}{Nature}} \textbf{\bibinfo{volume}{624}}, \bibinfo{pages}{570--578} (\bibinfo{year}{2023}).

\bibitem{swanson2025virtuallab}
\bibinfo{author}{Swanson, K.~, Wu, W.~, Bulaong, N.~L.~, Pak, J.~E.~ \& Zou, J.~}
\newblock \bibinfo{title}{The Virtual Lab of AI agents designs new SARS-CoV-2 nanobodies}.
\newblock \emph{\bibinfo{journal}{Nature}} \textbf{\bibinfo{volume}{646}}, \bibinfo{pages}{716--723} (\bibinfo{year}{2025}).

\bibitem{gottweis2026coscientist}
\bibinfo{author}{Gottweis, J.~ \emph{et~al.}}
\newblock \bibinfo{title}{Accelerating scientific discovery with Co-Scientist}.
\newblock \emph{\bibinfo{journal}{Nature}} \textbf{\bibinfo{volume}{655}}, \bibinfo{pages}{487--496} (\bibinfo{year}{2026}).

\bibitem{ghareeb2026robin}
\bibinfo{author}{Ghareeb, A.~E.~ \emph{et~al.}}
\newblock \bibinfo{title}{A multi-agent system for automating scientific discovery}.
\newblock \emph{\bibinfo{journal}{Nature}} \textbf{\bibinfo{volume}{655}}, \bibinfo{pages}{497--505} (\bibinfo{year}{2026}).

\bibitem{huang2026biomni}
\bibinfo{author}{Huang, K.~ \emph{et~al.}}
\newblock \bibinfo{title}{Autonomous biomedical research with an artificial intelligence agent}.
\newblock \emph{\bibinfo{journal}{Science}} \textbf{\bibinfo{volume}{393}}, \bibinfo{pages}{eadz4351} (\bibinfo{year}{2026}).

\bibitem{qiu2025biomars}
\bibinfo{author}{Qiu, Y.~ \emph{et~al.}}
\newblock \bibinfo{title}{BioMARS: A Multi-Agent Robotic System for Autonomous Biological Experiments}.
\newblock Preprint at \url{https://arxiv.org/abs/2507.01485v1} (2025).

\bibitem{eiraku2011opticcup}
\bibinfo{author}{Eiraku, M.~ \emph{et~al.}}
\newblock \bibinfo{title}{Self-organizing optic-cup morphogenesis in three-dimensional culture}.
\newblock \emph{\bibinfo{journal}{Nature}} \textbf{\bibinfo{volume}{472}}, \bibinfo{pages}{51--56} (\bibinfo{year}{2011}).

\bibitem{nakano2012retina}
\bibinfo{author}{Nakano, T.~ \emph{et~al.}}
\newblock \bibinfo{title}{Self-Formation of Optic Cups and Storable Stratified Neural Retina from Human ESCs}.
\newblock \emph{\bibinfo{journal}{Cell Stem Cell}} \textbf{\bibinfo{volume}{10}}, \bibinfo{pages}{771--785} (\bibinfo{year}{2012}).

\bibitem{zhong2014retinal}
\bibinfo{author}{Zhong, X.~ \emph{et~al.}}
\newblock \bibinfo{title}{Generation of three-dimensional retinal tissue with functional photoreceptors from human iPSCs}.
\newblock \emph{\bibinfo{journal}{Nature Communications}} \textbf{\bibinfo{volume}{5}}, \bibinfo{pages}{4047} (\bibinfo{year}{2014}).

\bibitem{cowan2020retina}
\bibinfo{author}{Cowan, C.~S.~ \emph{et~al.}}
\newblock \bibinfo{title}{Cell Types of the Human Retina and Its Organoids at Single-Cell Resolution}.
\newblock \emph{\bibinfo{journal}{Cell}} \textbf{\bibinfo{volume}{182}}, \bibinfo{pages}{1623--1640.e34} (\bibinfo{year}{2020}).

\bibitem{capowski2018staging}
\bibinfo{author}{Capowski, E.~E.~ \emph{et~al.}}
\newblock \bibinfo{title}{Reproducibility and staging of 3D human retinal organoids across multiple pluripotent stem cell lines}.
\newblock \emph{\bibinfo{journal}{Development}} \textbf{\bibinfo{volume}{146}}, \bibinfo{pages}{dev171686} (\bibinfo{year}{2019}).

\bibitem{brooks2019organoid}
\bibinfo{author}{Brooks, M.~J.~ \emph{et~al.}}
\newblock \bibinfo{title}{Improved Retinal Organoid Differentiation by Modulating Signaling Pathways Revealed by Comparative Transcriptome Analyses with Development In Vivo}.
\newblock \emph{\bibinfo{journal}{Stem Cell Reports}} \textbf{\bibinfo{volume}{13}}, \bibinfo{pages}{891--905} (\bibinfo{year}{2019}).

\bibitem{harkin2024organoid}
\bibinfo{author}{Harkin, J.~ \emph{et~al.}}
\newblock \bibinfo{title}{A highly reproducible and efficient method for retinal organoid differentiation from human pluripotent stem cells}.
\newblock \emph{\bibinfo{journal}{Proceedings of the National Academy of Sciences}} \textbf{\bibinfo{volume}{121}}, \bibinfo{pages}{e2317285121} (\bibinfo{year}{2024}).

\bibitem{lewis2020rag}
\bibinfo{author}{Lewis, P.~ \emph{et~al.}}
\newblock \bibinfo{title}{Retrieval-Augmented Generation for Knowledge-Intensive NLP Tasks}.
\newblock \emph{Advances in Neural Information Processing Systems} \textbf{33}, 9459--9474 (2020).

\bibitem{jin2025flashrag}
\bibinfo{author}{Jin, J.~ \emph{et~al.}}
\newblock \bibinfo{title}{FlashRAG: A Modular Toolkit for Efficient Retrieval-Augmented Generation Research}.
\newblock \emph{Companion Proceedings of the ACM on Web Conference 2025}, 737--740 (2025).

\bibitem{jiang2024protocode}
\bibinfo{author}{Jiang, S., Evans-Yamamoto, D., Bersenev, D., Palaniappan, S.~K. \& Yachie-Kinoshita, A.}
\newblock \bibinfo{title}{ProtoCode: Leveraging large language models (LLMs) for automated generation of machine-readable PCR protocols from scientific publications}.
\newblock \emph{SLAS Technology} \textbf{29}, 100134 (2024).

\bibitem{liu2024llambo}
\bibinfo{author}{Liu, T.~, Astorga, N.~, Seedat, N.~ \& van der Schaar, M.~}
\newblock \bibinfo{title}{Large Language Models to Enhance Bayesian Optimization}.
\newblock In \emph{The Twelfth International Conference on Learning Representations} (\bibinfo{year}{2024}).

\bibitem{jones1998ego}
\bibinfo{author}{Jones, D.~R.~, Schonlau, M.~ \& Welch, W.~J.~}
\newblock \bibinfo{title}{Efficient Global Optimization of Expensive Black-Box Functions}.
\newblock \emph{\bibinfo{journal}{Journal of Global Optimization}} \textbf{\bibinfo{volume}{13}}, \bibinfo{pages}{455--492} (\bibinfo{year}{1998}).

\bibitem{shahriari2016bo}
\bibinfo{author}{Shahriari, B.~, Swersky, K.~, Wang, Z.~, Adams, R.~P.~ \& de Freitas, N.~}
\newblock \bibinfo{title}{Taking the Human Out of the Loop: A Review of Bayesian Optimization}.
\newblock \emph{\bibinfo{journal}{Proceedings of the IEEE}} \textbf{\bibinfo{volume}{104}}, \bibinfo{pages}{148--175} (\bibinfo{year}{2016}).

\bibitem{frazier2018bo}
\bibinfo{author}{Frazier, P.~I.~}
\newblock \bibinfo{title}{Bayesian Optimization}.
\newblock \emph{\bibinfo{journal}{Recent Advances in Optimization and Modeling of Contemporary Problems}}, \bibinfo{pages}{255--278} (\bibinfo{year}{2018}).

\bibitem{latorre2016retinal}
\bibinfo{author}{La Torre, A.}
\newblock \bibinfo{title}{Retinal Differentiation of Mouse Embryonic Stem Cells}.
\newblock \emph{Bio-protocol} \textbf{6}, e1851 (2016).

\bibitem{himmelstein2017hetionet}
\bibinfo{author}{Himmelstein, D.~S.~ \emph{et~al.}}
\newblock \bibinfo{title}{Systematic integration of biomedical knowledge prioritizes drugs for repurposing}.
\newblock \emph{\bibinfo{journal}{eLife}} \textbf{\bibinfo{volume}{6}}, \bibinfo{pages}{e26726} (\bibinfo{year}{2017}).

\bibitem{chandak2023primekg}
\bibinfo{author}{Chandak, P.~, Huang, K.~ \& Zitnik, M.~}
\newblock \bibinfo{title}{Building a knowledge graph to enable precision medicine}.
\newblock \emph{\bibinfo{journal}{Scientific Data}} \textbf{\bibinfo{volume}{10}}, \bibinfo{pages}{67} (\bibinfo{year}{2023}).

\bibitem{gharizadeh2024hgtdr}
\bibinfo{author}{Gharizadeh, A.~, Abbasi, K.~, Ghareyazi, A.~, Mofrad, M.~R.~K.~ \& Rabiee, H.~R.~}
\newblock \bibinfo{title}{HGTDR: Advancing drug repurposing with heterogeneous graph transformers}.
\newblock \emph{\bibinfo{journal}{Bioinformatics}} \textbf{\bibinfo{volume}{40}}, \bibinfo{pages}{btae349} (\bibinfo{year}{2024}).

\bibitem{zitnik2018decagon}
\bibinfo{author}{Zitnik, M.~, Agrawal, M.~ \& Leskovec, J.~}
\newblock \bibinfo{title}{Modeling polypharmacy side effects with graph convolutional networks}.
\newblock \emph{\bibinfo{journal}{Bioinformatics}} \textbf{\bibinfo{volume}{34}}, \bibinfo{pages}{i457--i466} (\bibinfo{year}{2018}).

\bibitem{williams2001ifenprodil}
\bibinfo{author}{Williams, K.~}
\newblock \bibinfo{title}{Ifenprodil, a Novel NMDA Receptor Antagonist : Site and Mechanism of Action}.
\newblock \emph{\bibinfo{journal}{Current Drug Targets}} \textbf{\bibinfo{volume}{2}}, \bibinfo{pages}{285--298} (\bibinfo{year}{2001}).

\bibitem{gibson2002agmatine}
\bibinfo{author}{Gibson, D.~A.~, Harris, B.~R.~, Rogers, D.~T.~ \& Littleton, J.~M.~}
\newblock \bibinfo{title}{Radioligand binding studies reveal agmatine is a more selective antagonist for a polyamine-site on the NMDA receptor than arcaine or ifenprodil}.
\newblock \emph{\bibinfo{journal}{Brain Research}} \textbf{\bibinfo{volume}{952}}, \bibinfo{pages}{71--77} (\bibinfo{year}{2002}).

\bibitem{shuto2001evidence11325012}
\bibinfo{author}{Shuto, S., Yoshii, K. \& Matsuda, A.}
\newblock \bibinfo{title}{{(1S,2R)-1-Phenyl-2-[(S)-1-aminopropyl]-N,N-diethylcyclopropanecarboxamide (PPDC), a new class of NMDA-receptor antagonist: molecular design by a novel conformational restriction strategy}}.
\newblock \emph{\bibinfo{journal}{Jpn J Pharmacol}} \textbf{\bibinfo{volume}{85}}, \bibinfo{pages}{207--213} (\bibinfo{year}{2001}).

\bibitem{noguchi1999milnacipran}
\bibinfo{author}{Noguchi, T.~ \emph{et~al.}}
\newblock \bibinfo{title}{Open channel block of NMDA receptors by conformationally restricted analogs of milnacipran and their protective effect against NMDA-induced neurotoxicity}.
\newblock \emph{\bibinfo{journal}{Synapse}} \textbf{\bibinfo{volume}{31}}, \bibinfo{pages}{87--96} (\bibinfo{year}{1999}).

\bibitem{bacq2024stiripentol}
\bibinfo{author}{Bacq, A.~ \emph{et~al.}}
\newblock \bibinfo{title}{An Update on Stiripentol Mechanisms of Action: A Narrative Review}.
\newblock \emph{\bibinfo{journal}{Advances in Therapy}} \textbf{\bibinfo{volume}{41}}, \bibinfo{pages}{1351--1371} (\bibinfo{year}{2024}).

\bibitem{mantegazza2010phenytoin}
\bibinfo{author}{Mantegazza, M.~, Curia, G.~, Biagini, G.~, Ragsdale, D.~S.~ \& Avoli, M.~}
\newblock \bibinfo{title}{Voltage-gated sodium channels as therapeutic targets in epilepsy and other neurological disorders}.
\newblock \emph{\bibinfo{journal}{The Lancet Neurology}} \textbf{\bibinfo{volume}{9}}, \bibinfo{pages}{413--424} (\bibinfo{year}{2010}).

\bibitem{hardingham2010nmda}
\bibinfo{author}{Hardingham, G.~E.~ \& Bading, H.~}
\newblock \bibinfo{title}{Synaptic versus extrasynaptic NMDA receptor signalling: implications for neurodegenerative disorders}.
\newblock \emph{\bibinfo{journal}{Nature Reviews Neuroscience}} \textbf{\bibinfo{volume}{11}}, \bibinfo{pages}{682--696} (\bibinfo{year}{2010}).

\bibitem{papouin2012nmda}
\bibinfo{author}{Papouin, T.~ \emph{et~al.}}
\newblock \bibinfo{title}{Synaptic and Extrasynaptic NMDA Receptors Are Gated by Different Endogenous Coagonists}.
\newblock \emph{\bibinfo{journal}{Cell}} \textbf{\bibinfo{volume}{150}}, \bibinfo{pages}{633--646} (\bibinfo{year}{2012}).

\bibitem{eastlake2026muller}
\bibinfo{author}{Eastlake, K.~\emph{et~al.}}
\newblock \bibinfo{title}{M{\"u}ller glia derived EVs promote neurite recovery of an enriched population of retinal ganglion like cells derived from hESC retinal organoids after damage}.
\newblock \emph{\bibinfo{journal}{Scientific Reports}} \textbf{\bibinfo{volume}{16}}, \bibinfo{pages}{11853} (\bibinfo{year}{2026}).

\bibitem{steiner2019chemputer}
\bibinfo{author}{Steiner, S.~ \emph{et~al.}}
\newblock \bibinfo{title}{Organic synthesis in a modular robotic system driven by a chemical programming language}.
\newblock \emph{Science} \textbf{363}, eaav2211 (2019).

\bibitem{mehr2020digitization}
\bibinfo{author}{Mehr, S.~H.~M., Craven, M., Leonov, A.~I., Keenan, G. \& Cronin, L.}
\newblock \bibinfo{title}{A universal system for digitization and automatic execution of the chemical synthesis literature}.
\newblock \emph{Science} \textbf{370}, 101--108 (2020).

\bibitem{jocher2026yolo26}
\bibinfo{author}{Jocher, G.~ \emph{et~al.}}
\newblock \bibinfo{title}{Ultralytics YOLO26: Unified Real-Time End-to-End Vision Models}.
\newblock Preprint at \url{https://arxiv.org/abs/2606.03748} (2026).

\bibitem{hase2018phoenics}
\bibinfo{author}{Häse, F.~, Roch, L.~M.~, Kreisbeck, C.~ \& Aspuru-Guzik, A.~}
\newblock \bibinfo{title}{Phoenics: A Bayesian Optimizer for Chemistry}.
\newblock \emph{\bibinfo{journal}{ACS Central Science}} \textbf{\bibinfo{volume}{4}}, \bibinfo{pages}{1134--1145} (\bibinfo{year}{2018}).

\bibitem{shields2021reactionbo}
\bibinfo{author}{Shields, B.~J.~ \emph{et~al.}}
\newblock \bibinfo{title}{Bayesian reaction optimization as a tool for chemical synthesis}.
\newblock \emph{\bibinfo{journal}{Nature}} \textbf{\bibinfo{volume}{590}}, \bibinfo{pages}{89--96} (\bibinfo{year}{2021}).

\bibitem{internagent2025}
\bibinfo{author}{InternAgent Team}
\newblock \bibinfo{title}{InternAgent: When agent becomes the scientist---building closed-loop system from hypothesis to verification}.
\newblock Preprint at \url{https://arxiv.org/abs/2505.16938} (\bibinfo{year}{2025}).

\bibitem{rosarda2023imbalanced}
\bibinfo{author}{Rosarda, J.~D.~ \emph{et~al.}}
\newblock \bibinfo{title}{Imbalanced unfolded protein response signaling contributes to 1-deoxysphingolipid retinal toxicity}.
\newblock \emph{\bibinfo{journal}{Nature Communications}} \textbf{\bibinfo{volume}{14}}, \bibinfo{pages}{4119} (\bibinfo{year}{2023}).


\bibitem{mcclements2022tropism}
\bibinfo{author}{McClements, M.~E.~\emph{et~al.}}
\newblock \bibinfo{title}{Tropism of AAV Vectors in Photoreceptor-Like Cells of Human iPSC-Derived Retinal Organoids}.
\newblock \emph{\bibinfo{journal}{Translational Vision Science \& Technology}} \textbf{\bibinfo{volume}{11}}, \bibinfo{pages}{3} (\bibinfo{year}{2022}).

\bibitem{promega2023tm412}
\bibinfo{author}{Promega Corporation}.
\newblock \emph{\bibinfo{title}{CellTiter-Glo 3D Cell Viability Assay Technical Manual}}. TM412, Revised January 2023 (\bibinfo{year}{2023}).
\newblock \url{https://www.promega.com/resources/protocols/technical-manuals/101/celltiter-glo-3d-cell-viability-assay-protocol/}.

\bibitem{baptista2023polymer}
\bibinfo{author}{Baptista, D.~\emph{et~al.}}
\newblock \bibinfo{title}{Polymer film-based microwell array platform for long-term culture and research of human bronchial organoids}.
\newblock \emph{\bibinfo{journal}{Materials Today Bio}} \textbf{\bibinfo{volume}{19}}, \bibinfo{pages}{100603} (\bibinfo{year}{2023}).

\bibitem{vergara2017three}
\bibinfo{author}{Vergara, M.~N.~\emph{et~al.}}
\newblock \bibinfo{title}{Three-dimensional automated reporter quantification (3D-ARQ) technology enables quantitative screening in retinal organoids}.
\newblock \emph{\bibinfo{journal}{Development}} \textbf{\bibinfo{volume}{144}}, \bibinfo{pages}{3698--3705} (\bibinfo{year}{2017}).

\bibitem{fisher1936iris}
\bibinfo{author}{Fisher, R.~A.~}
\newblock \bibinfo{title}{The use of multiple measurements in taxonomic problems}.
\newblock \emph{\bibinfo{journal}{Annals of Eugenics}} \textbf{\bibinfo{volume}{7}}, \bibinfo{pages}{179--188} (\bibinfo{year}{1936}).

\bibitem{kew1998evidence9482779}
\bibinfo{author}{Kew, J.~N., Richards, J.~G., Mutel, V. \& Kemp, J.~A.}
\newblock \bibinfo{title}{{Developmental changes in NMDA receptor glycine affinity and ifenprodil sensitivity reveal three distinct populations of NMDA receptors in individual rat cortical neurons}}.
\newblock \emph{\bibinfo{journal}{J Neurosci}} \textbf{\bibinfo{volume}{18}}, \bibinfo{pages}{1935--1943} (\bibinfo{year}{1998}).

\bibitem{shuto1998evidence9719604}
\bibinfo{author}{Shuto, S., Ono, S., Imoto, H., Yoshii, K. \& Matsuda, A.}
\newblock \bibinfo{title}{{Synthesis and biological activity of conformationally restricted analogues of milnacipran: (1S, 2R)-1-phenyl-2-[(R)-1-amino-2-propynyl]-N,N- diethylcyclopropanecarboxamide is a novel class of NMDA receptor channel blocker}}.
\newblock \emph{\bibinfo{journal}{J Med Chem}} \textbf{\bibinfo{volume}{41}}, \bibinfo{pages}{3507--3514} (\bibinfo{year}{1998}).

\bibitem{xie2025evidence39969272}
\bibinfo{author}{Xie, T.~\emph{et~al.}}
\newblock \bibinfo{title}{{Agmatine inhibits NMDA receptor-mediated calcium transients in mouse spinal cord dorsal horn via intact PSD95-nNOS signaling}}.
\newblock \emph{\bibinfo{journal}{J Pharmacol Exp Ther}} \textbf{\bibinfo{volume}{392}}, \bibinfo{pages}{100061} (\bibinfo{year}{2025}).

\bibitem{azar2022agmatinenrf2}
\bibinfo{author}{Azar, Y.~O., Badawi, G.~A., Zaki, H.~F. \& Ibrahim, S.~M.}
\newblock \bibinfo{title}{{Agmatine-mediated inhibition of NMDA receptor expression and amelioration of dyskinesia via activation of Nrf2 and suppression of HMGB1/RAGE/TLR4/MYD88/NF-$\kappa$B signaling cascade in rotenone lesioned rats}}.
\newblock \emph{\bibinfo{journal}{Life Sciences}} \textbf{\bibinfo{volume}{311}}, \bibinfo{pages}{121049} (\bibinfo{year}{2022}).

\bibitem{monu2025agmatineretina}
\bibinfo{author}{Monu~\emph{et~al.}}
\newblock \bibinfo{title}{{Metabolomic Profiling of Aqueous Humor From Glaucoma Patients Identifies Metabolites With Anti-Inflammatory and Neuroprotective Potential in Mice}}.
\newblock \emph{\bibinfo{journal}{Investigative Ophthalmology \& Visual Science}} \textbf{\bibinfo{volume}{66}}, \bibinfo{pages}{28} (\bibinfo{year}{2025}).

\bibitem{verleye2016stiripentol}
\bibinfo{author}{Verleye, M., Buttigieg, D. \& Steinschneider, R.}
\newblock \bibinfo{title}{{Neuroprotective activity of stiripentol with a possible involvement of voltage-dependent calcium and sodium channels}}.
\newblock \emph{\bibinfo{journal}{Journal of Neuroscience Research}} \textbf{\bibinfo{volume}{94}}, \bibinfo{pages}{179--189} (\bibinfo{year}{2016}).

\bibitem{hirata2025evidence40651649}
\bibinfo{author}{Hirata, M., Kishimoto, Y., Shearer, T.~R. \& Azuma, M.}
\newblock \bibinfo{title}{{Hypoxia-induced activation of calpain and oxidative stress in mitochondria leads to RGC death in human iPSC-derived retinal organoids}}.
\newblock \emph{\bibinfo{journal}{Exp Eye Res}} \textbf{\bibinfo{volume}{259}}, \bibinfo{pages}{110511} (\bibinfo{year}{2025}).

\bibitem{yam2026evidence40277233}
\bibinfo{author}{Yam, M.~\emph{et~al.}}
\newblock \bibinfo{title}{{A mouse model of GRIN2D developmental and epileptic encephalopathy recapitulates the human disease}}.
\newblock \emph{\bibinfo{journal}{Brain}} \textbf{\bibinfo{volume}{149}}, \bibinfo{pages}{90--104} (\bibinfo{year}{2026}).

\bibitem{liu2024thsg}
\bibinfo{author}{Liu, T.~\emph{et~al.}}
\newblock \bibinfo{title}{{THSG alleviates cerebral ischemia/reperfusion injury via the GluN2B-CaMKII-ERK1/2 pathway}}.
\newblock \emph{\bibinfo{journal}{Phytomedicine}} \textbf{\bibinfo{volume}{129}}, \bibinfo{pages}{155595} (\bibinfo{year}{2024}).

\bibitem{kobayashi2017evidence28253402}
\bibinfo{author}{Kobayashi, M.~\emph{et~al.}}
\newblock \bibinfo{title}{{The Relationship Between the Renin-Angiotensin-Aldosterone System and NMDA Receptor-Mediated Signal and the Prevention of Retinal Ganglion Cell Death}}.
\newblock \emph{\bibinfo{journal}{Invest Ophthalmol Vis Sci}} \textbf{\bibinfo{volume}{58}}, \bibinfo{pages}{1397--1403} (\bibinfo{year}{2017}).

\bibitem{glasgow2017evidence28877967}
\bibinfo{author}{Glasgow, N.~G., Povysheva, N.~V., Azofeifa, A.~M. \& Johnson, J.~W.}
\newblock \bibinfo{title}{{Memantine and Ketamine Differentially Alter NMDA Receptor Desensitization}}.
\newblock \emph{\bibinfo{journal}{J Neurosci}} \textbf{\bibinfo{volume}{37}}, \bibinfo{pages}{9686--9704} (\bibinfo{year}{2017}).

\bibitem{lee2023evidence37641807}
\bibinfo{author}{Lee, J.~R. \& Jeong, K.~W.}
\newblock \bibinfo{title}{{N-retinylidene-N-retinylethanolamine degradation in human retinal pigment epithelial cells via memantine- and ifenprodil-mediated autophagy}}.
\newblock \emph{\bibinfo{journal}{Korean J Physiol Pharmacol}} \textbf{\bibinfo{volume}{27}}, \bibinfo{pages}{449--456} (\bibinfo{year}{2023}).

\bibitem{kong2024evidence38685368}
\bibinfo{author}{Kong, C.~H.~\emph{et~al.}}
\newblock \bibinfo{title}{{Cheonwangbosimdan mitigates post-traumatic stress disorder-like behaviors through GluN2B-containing NMDA receptor antagonism in mice}}.
\newblock \emph{\bibinfo{journal}{J Ethnopharmacol}} \textbf{\bibinfo{volume}{330}}, \bibinfo{pages}{118270} (\bibinfo{year}{2024}).

\bibitem{kotagale2019evidence31063707}
\bibinfo{author}{Kotagale, N.~R., Taksande, B.~G. \& Inamdar, N.~N.}
\newblock \bibinfo{title}{{Neuroprotective offerings by agmatine}}.
\newblock \emph{\bibinfo{journal}{Neurotoxicology}} \textbf{\bibinfo{volume}{73}}, \bibinfo{pages}{228--245} (\bibinfo{year}{2019}).

\bibitem{zhang2026evidence41811034}
\bibinfo{author}{Zhang, Z.~\emph{et~al.}}
\newblock \bibinfo{title}{{Discovery and Evaluation of N-Arylindole-Based GluN2B-NMDAR Antagonists with Reduced Cardiotoxicity for the Treatment of Ischemic Stroke}}.
\newblock \emph{\bibinfo{journal}{J Med Chem}} \textbf{\bibinfo{volume}{69}}, \bibinfo{pages}{7290--7312} (\bibinfo{year}{2026}).

\bibitem{haque2021evidence33285471}
\bibinfo{author}{Haque, M.~N., Hannan, M.~A., Dash, R., Choi, S.~M. \& Moon, I.~S.}
\newblock \bibinfo{title}{{The potential LXR$\beta$ agonist stigmasterol protects against hypoxia/reoxygenation injury by modulating mitophagy in primary hippocampal neurons}}.
\newblock \emph{\bibinfo{journal}{Phytomedicine}} \textbf{\bibinfo{volume}{81}}, \bibinfo{pages}{153415} (\bibinfo{year}{2021}).

\bibitem{iwata2022evidence35435002}
\bibinfo{author}{Iwata, Y.~\emph{et~al.}}
\newblock \bibinfo{title}{{Protective effect of d-alanine against acute kidney injury}}.
\newblock \emph{\bibinfo{journal}{Am J Physiol Renal Physiol}} \textbf{\bibinfo{volume}{322}}, \bibinfo{pages}{F667--F679} (\bibinfo{year}{2022}).

\end{thebibliography}
